\documentclass[journal]{IEEEtran}
\usepackage{amsmath,amsfonts}
\usepackage{algorithmic}
\usepackage{algorithm}

\usepackage{array}
\usepackage[caption=false,font=normalsize,labelfont=sf,textfont=sf]{subfig}
\usepackage{textcomp}
\usepackage{stfloats}
\usepackage{url}
\usepackage{verbatim}
\usepackage{graphicx}
\usepackage{cite}
\usepackage{amsthm,amsthm}
\usepackage{amssymb}
\usepackage{multirow}
\usepackage{xcolor}
\usepackage{marginnote}
\usepackage{etoolbox}   
\usepackage{ifoddpage}

\makeatother
\newtheorem{lemma}{Lemma}

\newtheorem{Def}{Definition}
\newtheorem{rmk}{Remark}

\usepackage{flushend}
\usepackage{caption}  
\begin{document}
\title{An Efficient Real-Time Planning Method for Swarm Robotics Based on an Optimal Virtual Tube}

\author{Pengda Mao, Shuli Lv, Chen Min, Zhaolong Shen, and Quan Quan,~\IEEEmembership{Senior Member,~IEEE}
	\thanks{Pengda Mao, Shuli Lv, Chen Min, Zhaolong Shen and Quan Quan are with School of Automation Science and Electrical Engineering,
		Beihang University, Beijing, 100191, P.R. China
		{\tt\small \{maopengda, lvshuli, min\_chen, shenzhaolong, qq\_buaa\}@buaa.edu.cn}}}



\maketitle

\begin{abstract}	
 \textcolor{blue}{Robot swarms} navigating through unknown obstacle environments is an emerging research area that faces challenges. 
 Performing tasks in such environments requires swarms to achieve autonomous localization, perception, decision-making, control, and planning. The limited computational resources of onboard platforms present significant challenges for planning and control.
 Reactive planners offer low computational demands and high re-planning frequencies but lack predictive capabilities, often resulting in local minima. \textcolor{blue}{Multi-step planners can make multi-step predictions to reduce deadlocks, but they require substantial computation, which leads to lower replanning frequency.
This paper proposes a novel homotopic trajectory planning framework for robot swarm that combines centralized homotopic trajectory planning (optimal virtual tube planning) with distributed control, enabling low-computation, high-frequency replanning, thereby uniting the strengths of multi-step and reactive planners. 
 Based on multi-parametric programming, homotopic optimal trajectories are approximated by affine functions. 
 The resulting approximate solutions have computational complexity $O(n_t)$, where $n_t$ is the number of trajectory parameters. 
 This low complexity makes centralized planning of a large number of optimal trajectories practical and, when combined with distributed control, enables rapid, low-cost replanning.}
 The effectiveness of the proposed method is validated through several simulations and experiments.
\end{abstract}


\section{Introduction}

\textcolor{blue}{Safely and effectively navigating obstacle-dense environments is a central challenge for robot swarms. This capability is crucial for applications such as search-and-rescue, environmental monitoring, and infrastructure inspection, where robots must coordinate under limited sensing, communication, and computational resources to avoid collisions and complete mission objectives.}

\textcolor{blue}{Robot swarm trajectory planning is often classified into three categories: centralized, distributed, and decentralized.} 
In centralized trajectory planning, the leader robot receives the state information of all robots within the swarm, plans collision avoidance trajectories for all robots, and then distributes the trajectories for each robot to execute trajectory tracking control. Centralized methods are straightforward to deploy, can reduce localization errors, and keep per-robot hardware requirements low; however, they face scalability challenges in communication and computation as the swarm size grows.
\textcolor{blue}{In distributed/decentralized trajectory planning, each robot independently plans its own trajectory to achieve collision avoidance and shares its planned trajectory (or local state) with neighboring robots via communication. This enables rapid local responses to unexpected events. Based on prediction horizon length, such local planners can be further categorized into reactive planners and multi-step planners \cite{csenbacslar2022asynchronous}.}

Reactive planners, such as the potential field method\textcolor{blue}{\cite{hwang1992potential}} and control barrier function (CBF)\textcolor{blue}{\cite{ames2019control}}, plan the next motion based on the current states of robots in neighbors. The advantages of reactive planners include low computational complexity, high replanning frequency, and minimal impact from communication delays. However, due to the lack of prediction steps, the reactive planners are prone to getting trapped in local minima and may exhibit less smooth movements.
\textcolor{blue}{Multi-step planners, on the other hand, predict multiple steps.
These multi-step actions may be represented either in discrete form \cite{toumieh2022decentralized} or as continuous form \cite{tordesillas2021mader} (e.g., parameterized trajectories).}
The advantages of multi-step planners include smoother movements and reduced deadlock due to the prediction horizon. However, the drawbacks of the multi-step planners are also evident, as trajectory planning involves substantial computation costs and is sensitive to communication delays.

\begin{figure}
	\centering
	\includegraphics[width=\linewidth]{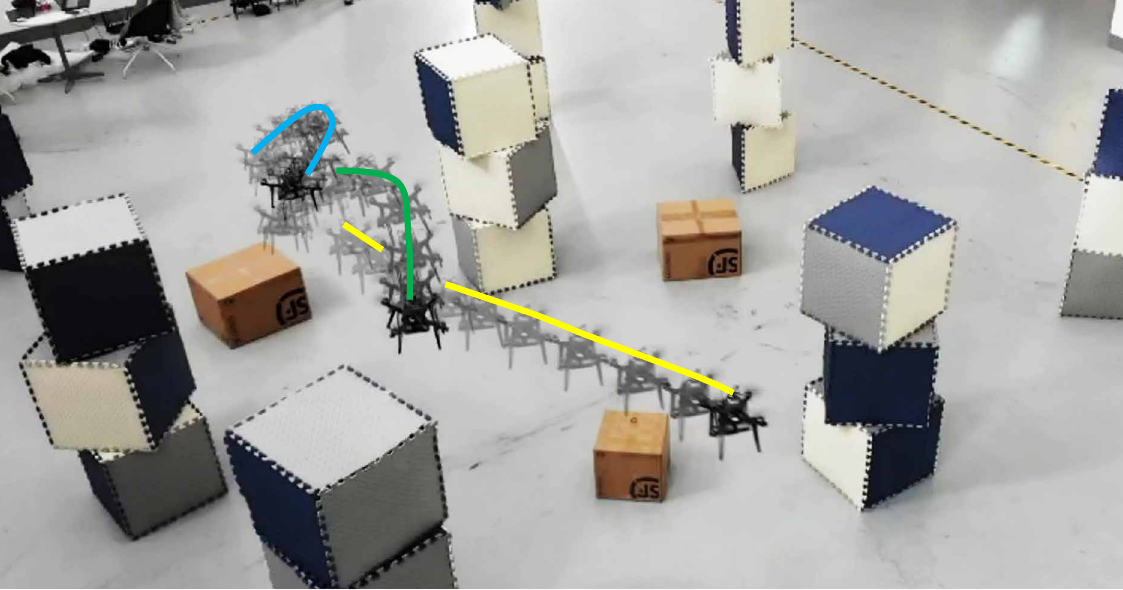}
	\caption{\textcolor{blue}{Overlaid trajectories of three drones navigating an unknown obstacle environments (proposed method). Trajectories are shown as temporal motion trails and color-coded in yellow, green, and blue.}}
	\label{fig:head-pic}
\end{figure}

\textcolor{blue}{Recently, homotopic trajectory planning (optimal virtual tube planning) \cite{mao2023optimal,mao2024tube,deng2025bioinspired,huang2024homotopic} has emerged as an efficient approach for swarm trajectory planning with certain collective properties.The core idea is to treat all trajectories of the swarm as a set and to plan over a set of homotopic trajectories instead of a single trajectory, thereby enabling more efficient computation. Homotopic trajectory planning comprises two steps: first, planning homotopic paths \cite{mao2024tube,huang2024homotopic}; second, generating homotopic trajectories based on those paths \cite{mao2023optimal}. However, existing methods suffer from the following shortcomings:
	\begin{itemize}
		\item \textbf{Not real-time}: Although the efficiency of homotopic optimal trajectories has been demonstrated, real-time generation in unknown environments has not been realized.
		\item \textbf{Lack of generality}: current works address only spatial optimization for piece-wise continuous polynomial trajectories and does not solve the spatial-temporal optimization problem for other parameterized trajectories with strong nonlinearities.
	\end{itemize}
To resolve these issues, this paper proposes a real-time homotopic-trajectory planning framework for unknown environments. The framework embeds homotopic path and trajectory planning within a hierarchical architecture and incorporates multi-parametric programming to address the spatial-temporal optimization of parameterized homotopic optimal trajectories.
Multi-parametric programming approximates optimal trajectories with affine functions; because these admit very fast evaluation, the proposed method enables rapid replanning and thus provides the responsiveness and smoothness required for real-time swarm motion.
The main contributions of this paper are summarized as follows:
}


\begin{itemize}
	\item A novel real-time planning framework for homotopic trajectory planning (optimal virtual tube planning) is proposed that combines centralized trajectory planning with distributed control. Specifically, one robot in the swarm plans an optimal virtual tube and shares it with all robots in the swarm, achieving centralized planning for trajectories. The task of collision avoidance is delegated to distributed control, enabling rapid responses to unexpected situations. 
\textcolor{blue}{\item The homotopic trajectory planning is generalized to a spatial-temporal formulation. Leveraging multi-parametric programming, the core optimization-solving process is independent of the number of trajectories to be generated, since only a fixed number of optimization problems need to be solved to obtain the optimal boundary trajectories of the virtual tube. The remaining trajectories can be generated efficiently through affine mappings, with computational complexity $O(n_t)$ in the number of optimization variables $n_t$. This makes the method particularly suitable for large swarm trajectory planning.
\item Numerical simulations validate the proposed method and analyze parameter effects. Comparative results demonstrate the computational efficiency, and several experiments in unknown obstacle environments confirm its real-time performance and effectiveness.}

\end{itemize}
 
\section{Related Work}
This section will introduce the state-of-the-art methods, including reactive planners and \textcolor{blue}{multi-step} planners, in the trajectory planning of \textcolor{blue}{robot swarms}, pointing out their respective advantages and disadvantages, and then discuss the current research on the virtual tube for \textcolor{blue}{robot swarms}.
\subsection{Reactive Planner}
Reactive planners determine the commands for the next moment based on the current state. These methods include artificial potential field (APF), control barrier function (CBF), bio-inspired methods, \textcolor{blue}{dynamic window approach (DWA), Voronoi tessellation method, velocity obstacle (VO) paradigm, and formation control}.
The artificial potential field (APF) method \cite{khatib1986real} is a traditional and effective method. Inspired by the concepts of electric and magnetic fields in physics, it translates the attraction to the goal and repulsion from obstacles or other robots into a mathematical model to guide robots from initial positions to target positions. Its advantages include simplicity, stability, and ease of implementation. However, it also faces challenges such as local minima \cite{hernandez2011convergence} and nonsmooth commands. Many subsequent studies have focused on optimizing the potential field function and avoiding local minima \cite{rostami2019obstacle,antich2005extending,ge2005queues}.
The CBF methods \cite{ames2019control} enforce collision avoidance by adding state constraints into an optimization framework, making them suitable for precise control and complex-constraint scenarios; they yield smoother control inputs than artificial potential fields (APF) but at higher computational cost, and APF has been shown to be a special case of CBF for obstacle avoidance \cite{singletary2021comparative}. In robotic swarms, CBFs can simultaneously achieve control objectives, collision avoidance, and connectivity maintenance while ensuring QP feasibility \cite{wang2016multi}, and have been extended to nonsmooth barrier functions to handle collisions both within the swarm and with obstacles \cite{glotfelter2017nonsmooth}.
Bio-inspired methods mimic collective behaviors—avoid collisions, align velocities, and move toward the group centroid—to produce simple, cohesive swarm motion, with classic models like Vicsek \cite{vasarhelyi2018optimized} and Boids \cite{reynolds1987flocks}; however, they offer no formal safety guarantees and are sensitive to parameter choices. Recently, optimization-based approaches have leveraged principles of Boids to define local cost functions, enabling distributed, constraint-aware control that improves efficiency and robustness. \cite{olfati2006flocking, beaver2020optimal, liu2021hierarchical}
\textcolor{blue}{The DWA samples feasible velocity commands within the dynamic limits of robots, plans next step for each, evaluates them by the gain function, and selects the best command for collision-free, dynamically-feasible motion \cite{fox2002dynamic}. \cite{brock1999high} extends the model to non-holonomic models in the unknown environment, while \cite{ogren2005convergent} attempts to resolve the deadlock problem by addressing the convergence properties of the method.
The Voronoi tessellation method partitions a space into Voronoi cells where each point in a cell is closer to its defining seed point than to any other, and it is applied in coverage control \cite{cortes2004coverage} and path planning \cite{gao2025swarmcvt}.
Many studies have improved upon the Voronoi diagram, including real-time updates of the Voronoi diagram \cite{kemna2017multi}, consideration of noise within the swarm \cite{zhou2017fast}, and discrete Voronoi diagram for large-scale swarm systems \cite{han2023hybrid}. However, generating a Voronoi diagram in three-dimensional space requires significant computational effort, necessitating more efficient algorithms \cite{zhou2024multi}.
The Velocity Obstacle (VO) paradigm \cite{fiorini1998motion} is a geometric collision-avoidance method that, for each moving obstacle, computes the set of robot velocities that would lead to a collision in next step and then chooses a velocity outside the union of these forbidden sets to ensure collision-free motion. Reciprocal velocity obstacle (RVO) \cite{van2008reciprocal} improve the sub-optimal collision avoidance by sharing the responsibility between robots. Optimal reciprocal collision avoidance (ORCA) \cite{van2011reciprocal} selects new velocity by  solving a constrained linear optimization problem to guarantee optimality, smooth and safe motions.
Formation control enables robot swarm to maintain or change their relative positions to form and preserve a desired geometric configuration for coordinated task execution. The existing formation control approaches can be classified into displacement-based, distance-based, and bearing-based approaches \cite{oh2015survey,zhao2015bearing,zhao2015translational}. Formation control can enable a swarm to pass through the obstacle environments by changing its formation, but the formations often lack sufficient flexibility \cite{zhao2018affine}.
Overall, because reactive planners only need to predict the next action, they are computationally efficient; however, this limited foresight can cause oscillations and entrapment in local minima. When navigating complex environments, they should be used with a global planner.}
%
\subsection{\textcolor{blue}{Multi-step} Planner}
The \textcolor{blue}{multi-step} planner plans multi-step predictions for the robot based on the current states of the swarm and the shared multi-step predictions of adjacent robots. Compared to the reactive planner, it ensures smoother motion and reduces the likelihood of deadlock. However, it has a lower replanning frequency, higher computational cost, and is affected by communication delays. The multi-step predictions of the \textcolor{blue}{multi-step} planner can be represented in either discrete \cite{toumieh2022decentralized} or continuous forms \cite{tordesillas2021mader}.
A typical method of discrete forms is multi-agent Model Predictive Control (MPC). Unlike single-agent MPC, multi-agent MPC not only incorporates discrete models and obstacle avoidance constraints but also uses the shared discrete trajectories of other robots as collision avoidance constraints. This method is suitable for trajectory planning in complex, high-order systems, but the strong nonlinearity and numerous variables involved result in higher computational costs. Many studies have sought to improve computational efficiency for real-time planning by linearizing non-convex constraints and adding relaxation variables \cite{soria2021distributed,luis2020online}.
In contrast, the continuous form of \textcolor{blue}{multi-step} planners usually employs parameterized trajectories, such as B-spline curves, to represent continuous multi-step predictions. This method helps mitigate issues related to excessive variables and lack of global information. However, trajectory planning methods also face complex non-convex collision avoidance constraints and are unsuitable for precise control of complex systems due to the lack of consideration for the system model. Thus, many studies focus on simplifying these constraints. Some have introduced safe flight corridors (SFC) \cite{Park2021Online}, used in single-robot planning, into multi-robot planning to simplify the representation of collision avoidance constraints. Others have transformed constrained optimization problems into unconstrained ones using penalty functions to accelerate computation \cite{zhou_ego-swarm_2021}. Additionally, some studies have addressed the effect of communication delays by setting different planning states \cite{kondo2023robust}.
\subsection{Virtual Tube}
The ``virtual tube" concept serves as a crucial safety area, delineating both the motion boundaries and directions for swarms. Originally conceived in the aerial traffic \cite{airbus} to ensure safe airspace for unmanned aerial vehicles, these virtual tubes establish safeguarded aerial space, preventing conflicts with ground or airborne traffic. Building upon this foundation, Quan et al. \cite{Quan2021Practical} extended this concept, originally applied to single-vehicle flight, to the domain of \textcolor{blue}{robot swarms}. This virtual tube, devoid of internal obstacles, functions as a safety area and simplifies the control, which only needs to avoid inter-robot collisions and ensure non-contact with the boundary of the virtual tube. 
Moreover, expanding the applications to guiding swarms through passages, apertures, and around surveillance targets without divergence \cite{gao2022multi,guo2024distributed}, the virtual tube is extended to encompass the concept of ``curved virtual tube" \cite{Quan2021Distributed}. This curved virtual tube resembles road lanes for autonomous vehicles \cite{rasekhipour2016potential}, \cite{luo2018porca}, and corridors within multi-UAV frameworks \cite{nagrare2022multi}.

In summary, the problem of the virtual tube can be divided into two core challenges—the \emph{virtual tube planning problem} and the \emph{virtual-tube-based passing-through control problem}. It should be noted that the methods mentioned above primarily address the passing-through control problem.
For the virtual tube planning problem, the recent works \cite{Mao2022,rao2024gradient} outline a trajectory planning method yielding a generator curve through discrete waypoints via search-based methods. Subsequently, the virtual tube emerges via the expansion of this generator curve, meticulously avoiding obstacles. To plan the virtual tube in the high-dimension space, a virtual tube without a generator curve is defined, and an optimal virtual tube is proposed to generate infinite optimal trajectories \cite{mao2023optimal,mao2024tube}. 
However, an efficient planning framework is needed to achieve real-time planning of the optimal virtual tube in unknown obstacle environments. 
Furthermore, previous works focused on specific piecewise polynomial trajectories and assumed consistent time allocations. Thus, there is a need to generalize the planning method to accommodate different trajectory representations and objectives.

\section{Preliminaries and Problem Formulation}
\textcolor{blue}{This section presents the necessary background for swarm trajectory planning. It covers the parameterization of trajectories using B\'{e}zier curves, the definition of the virtual tube (as introduced in \cite{mao2023optimal}), modeling assumptions (point-mass robots), and relevant problem constraints—thereby laying the foundation for the proposed planning methods.}
\subsection{Trajectory Parameterization}
\textcolor{blue}{It is common to employ a B\'ezier curve of degree $p$ to represent the trajectory \cite{gao2019flying,chen2017accurate,durakli2022new}; exploiting the convex-hull property, the curve lies entirely within the convex hull of its control points. B\'ezier curves are typically used in $\mathbb{R}^2$ or $\mathbb{R}^3$, and the representation in $\mathbb{R}^3$ is described in the following \cite{sederberg2012computer}.}
Let the set of control points be ${\bf{P}} = \left[ 
		{{{\bf{p}}_0}}\quad{{{\bf{p}}_1}}\quad\cdots \quad{{{\bf{p}}_p}} \right] \in {\mathbb{R}^{3 \times (p + 1)}}$ and the domain be $t \in \left[ {0,\Delta t} \right]$.
The B\'{e}zier curve is represented by
\begin{equation}
{\bf{h}}\left( {t} \right) = \sum\nolimits_{k = 0}^p {{B_{p,k}}\left( t \right)} {{\bf{p}}_k},
\label{equ:b-spline}
\end{equation}
where ${\bf{h}}\left( 0 \right) = {{\bf{p}}_0}$, ${\bf{h}}\left( \Delta t \right) = {{\bf{p}}_p}$ and the basis functions ${B_{p,k}}\left(t\right)$ are expressed as
\[{B_{p,k}}\left( t \right) = \frac{{p!}}{{k!\left( {p - k} \right)!}}{\left( {\frac{t}{{\Delta t}}} \right)^k}{\left( {1 - \frac{t}{{\Delta t}}} \right)^{p - k}}.\]
And the derivative of the B\'{e}zier curve in (\ref{equ:b-spline}) is expressed as
\begin{equation}
\dot{\bf{ h}}\left( t \right) = \sum\nolimits_{k = 0}^{p - 1} {{B_{p - 1,k}}\left( t \right)} {{\bf{q}}_k}, 
\end{equation}
where ${{{\bf{q}}_k}}$ are defined as follows:
\begin{equation}
{{\bf{q}}_k} = \frac{p}{{\Delta t}}\left( {{{\bf{p}}_{k + 1}} - {{\bf{p}}_k}} \right).
\end{equation}
Similarly, the derivative of the B\'{e}zier curve is bounded by the control points ${\bf q}_k$, which is useful for constraining high-order systems. 

The representation of the B\'{e}zier curve in matrix form is useful for the optimization problem. While there is a non-standard representation, specific degrees of B\'{e}zier curves have been derived, as detailed in \cite{joy2000matrix}. \textcolor{blue}{As shown in \cite{joy2000matrix},} a cubic B\'{e}zier curve can be written as
\begin{equation}
	{\bf{h}}\left( t \right) = \boldsymbol{\beta} \left( t \right){{\bf{S}}_{\Delta t}}{\bf{M}}{{\bf{P}}^{\rm{T}}},
	\label{equ:bezier-mat}
\end{equation}
where $\boldsymbol{\beta} \left( t \right) = [1\ t\ {t^2}\ {t^3}]$, 
\[{{\bf{S}}_{\Delta t}} = \left[ {\begin{array}{*{20}{c}}
		1&0&0&0\\
		0&{\frac{1}{{\Delta t}}}&0&0\\
		0&0&{\frac{1}{{\Delta {t^2}}}}&0\\
		0&0&0&{\frac{1}{{\Delta {t^3}}}}
\end{array}} \right],\]
\[{\bf{M}} = \left[ {\begin{array}{*{20}{c}}
		1&0&0&0\\
		{ - 3}&3&0&0\\
		3&{ - 6}&3&0\\
		{ - 1}&3&{ - 3}&1
\end{array}} \right].\]
\subsection{Virtual Tube}
\begin{figure}
	\centering
	\includegraphics{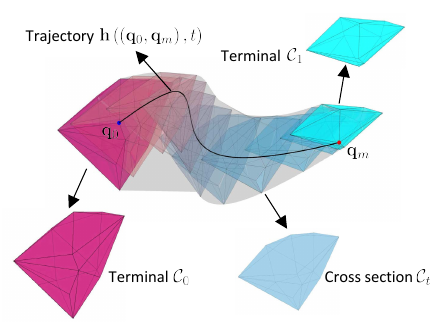}
	\caption{An example of a virtual tube (\textcolor{blue}{modified from \cite{mao2023optimal}}). The purple and blue polyhedrons are terminals. The shaded polyhedrons are cross sections ${\mathcal C}_t$. The black curve is a trajectory that is from ${\bf q}_{0}$ in terminal $\mathcal{C}_0$ to ${\bf q}_{m}$ in terminal $\mathcal{C}_1$. The gray area is the virtual tube.}
	\label{fig:virtual-tube}
\end{figure}

\textcolor{blue}{A virtual tube, defined by maps between bounded convex sets in free space, provides an \textbf{infinite set of homotopic trajectories}. This is particularly useful for guiding a robotic swarm through an environment as a cohesive group, as it ensures topological consistency (i.e., paths can be continuously deformed into one another without crossing obstacles) while allowing individual robots flexibility within the tube.}
\textcolor{blue}{The rigorous mathematical definition of the virtual tube was first established in \cite{mao2023optimal}. 
For completeness and to provide context, the definition of the \emph{virtual tube} is presented first, followed by the definition of \emph{homotopic paths} that underlie it. 	
}
\begin{Def}[\textcolor{blue}{Virtual Tube\cite{mao2023optimal}}]
 	A \emph{virtual tube} $\mathcal T$, as shown in Fig. \ref{fig:virtual-tube}, is a set in $n$-dimension space represented by a 4-tuple $\left( {\mathcal C}_0, {\mathcal C}_1, {\bf f}, {\bf h} \right)$ where
 	\begin{itemize}
 		\item ${\mathcal C}_0, {\mathcal C}_1$, called \emph{terminals}, are disjoint bounded convex subsets in $n$-dimension space.
 		\item ${\bf f}$ is a \emph{diffeomorphism}: ${\mathcal C}_0 \to {\mathcal C}_1$, so that there is a set of order pairs ${\mathcal{P}} = \left\{ {\left( {{\bf q}_0,{\bf q}_m} \right)| {\bf q}_0 \in {\mathcal{C}_0},{\bf q}_m = {\bf f}\left( {\bf q}_0 \right) \in {\mathcal{C}}_1} \right\}$.
 		\item ${\bf h}$ is a smooth\footnote{A real-valued function is said to be smooth if its derivatives of all orders exist and are continuous.} map: ${\mathcal P} \times {\mathcal I} \to {\mathcal T} $ where ${\mathcal I} = [0,1]$, such that ${\mathcal T} = \{ {\bf h} \left( \left({\bf q}_0,{\bf q}_m\right),t\right) | \left({\bf q}_0,{\bf q}_m\right) \in {\mathcal P} , t \in {\mathcal I} \}$, ${\bf h}\left( {\left( {\bf q}_0,{\bf q}_m \right),0} \right) = {\bf q}_0$, ${\bf h}\left( {\left( {\bf q}_0,{\bf q}_m \right),1} \right) = {\bf q}_m$. The function ${\bf h}\left( {\left( {\bf q}_0,{\bf q}_m \right),t} \right)$ is called a \emph{trajectory} for a order pair $\left( {\bf q}_0,{\bf q}_m\right)$.
 	\end{itemize}
 	And, a \emph{cross-section} ${\mathcal{C}}$ of a virtual tube at $t \in {\mathcal{I}}$ is expressed as:
 	\begin{equation}
 	{\mathcal{C}}_{t} = \left\{ {{\bf h}\left( {\left( {\bf q}_0,{\bf q}_m \right),{t}} \right)|\left({\bf q}_0,{\bf q}_m\right) \in {\mathcal{P}} } \right\}.
 	\end{equation}
 	The surface of the virtual tube is the boundary of ${\mathcal{T}}$, defined as $\partial {\mathcal{T}}$.
 	\label{def:virtual-tube}
\end{Def}

And if each trajectory in the virtual tube is optimal, then the virtual tube is optimal. The definition of optimal virtual tube is as follows.
\begin{Def}[\textcolor{blue}{Optimal Virtual Tube\cite{mao2023optimal}}]
	A virtual tube $\mathcal T^*$ is \emph{optimal} with respect to a cost  $g$ if every trajectory ${\bf h}^*$ in the tube is \emph{optimal} with respect to a cost $g$, namely
	${\cal T}^* = \left( {\mathcal{C}_0,\mathcal{C}_1,{\bf f},{{\bf h}^*}} \right).$
\end{Def}
\textcolor{blue}{
\begin{Def}[Path\cite{mao2024tube}]
	For any order pair $\left( {{{\bf{q}}_{0}},{{\bf{q}}_{m}}} \right)\in \mathcal{P}$, the \emph{path} is a continuous function $\boldsymbol{\sigma}\left(\left({\bf q}_0,{\bf q}_m\right),t\right) :{\mathcal P} \times {\mathcal I} \to { X}$ where $\boldsymbol{\sigma}\left(\left({\bf q}_0,{\bf q}_m\right),0\right)={\bf q}_0$ and $\boldsymbol{\sigma}\left(\left({\bf q}_0,{\bf q}_m\right),1\right)={\bf q}_m$, $m$ represents the number of the path points. And, the paths $\boldsymbol{\sigma}\left(\left({\bf q}_{0,k},{\bf q}_{m,k}\right),t\right)$ is the boundary paths if the points ${\bf q}_{0,k}$ and ${\bf q}_{m,k}$ are the vertexes of the terminals $\mathcal{C}_0$ and $\mathcal{C}_1$ respectively.
\end{Def}
\begin{Def}[Homotopic Paths\cite{mao2024tube}]
	The paths ${\boldsymbol{{\sigma}}}_1$ and ${\boldsymbol{{\sigma}}}_2$ are \emph{homotopic} if there is a continuous map ${\bf H}:\mathcal{I} \times \mathcal{I}\to X$ such that ${\bf{H}}\left( {t,0} \right) = {\boldsymbol{\sigma} _1}\left( {\left( {{{\bf{q}}_{0,1}},{{\bf{q}}_{m,1}}} \right),t} \right)$ $,{\bf{H}}\left( {t,1} \right) = {\boldsymbol{\sigma} _2}\left( {\left( {{{\bf{q}}_{0,2}},{{\bf{q}}_{m,2}}} \right),t} \right)$, ${\bf{H}}\left( {0,0} \right) = {{\bf{q}}_{0,1}}$, ${\bf{H}}\left( {1,0} \right) = {{\bf{q}}_{m,1}}$, ${\bf{H}}\left( {0,1} \right) = {{\bf{q}}_{0,2}}$$,{\bf{H}}\left( {1,1} \right) = {{\bf{q}}_{m,2}}$.
	{Moreover, consider a set of paths $ \Sigma_{\boldsymbol{\sigma}} $. The $ \Sigma_{\boldsymbol{\sigma}} $ is in the same homotopy class, if any ${\boldsymbol{{\sigma}}}_1,\boldsymbol{{\sigma}}_2 \in \Sigma_{\boldsymbol{\sigma}}$ are homotopic. For convenience, this $ \Sigma_{\boldsymbol{\sigma}} $ is called \emph{homotopic paths}. }
	\label{def:homo-paths}
\end{Def}
}
\subsection{Robot Model}
The robot in swarm is regarded as a mass point which is expressed as
\begin{equation}
\dot{{\bf{ p}}_i} = {{\bf{v}}_{c,i}},
\label{equ:robotmodel}
\end{equation}
where ${\bf p}_i$ is the position of the $i$-th robot and ${\bf v}_{c,i} $ is the velocity command for the $i$-th robot in swarm. According to \cite{Quan2021Practical}, there are three areas around the robot: safety, avoidance, and other areas. The safety area is a sphere with a radius $r_\text{s}$ centered into the center of mass of the robot to represent the physical shape of the robot. If other robots enter the safety area, a collision may happen. Thus, other robots are not allowed to enter the safety area. And the avoidance area is a sphere with a radius $r_\text{a}$ centered in the center of mass of the robot to avoid other robots. When the avoidance area has intersection with safety areas of other robots, avoidance motions are triggered. When other robots are in the other area, there is no avoidance between them.
\begin{figure}
	\centering
	\includegraphics{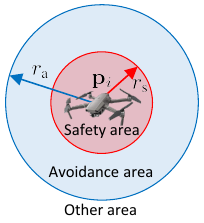}
	\caption{The robot model. The red and blue areas represent the safety area with a radius $r_\text{s}$ and the avoidance area with a radius $r_\text{a}$, respectively.}
	\label{fig:robot-model}
\end{figure}

\subsection{Multi-Parametric Programming}
Multi-parametric programming is a paradigm that considers the change of solution of a constrained optimization problem as a function of parameters.
Multi-parametric programming detaches from the sensitivity analysis theory \cite{fiacco1983introduction}. The process conditions of sensitivity analysis deviate from the nominal values to its neighborhood; multi-parametric programming is concerned with the whole range of the parameters.
\textcolor{blue}{
A general form \cite{pistikopoulos2007multi} of the multi-parametric programming is defined as
\begin{equation}
	\begin{array}{rl}
		V\left( {\boldsymbol{\theta }} \right) = \mathop {\min }\limits_{\bf{x}}  & {f_0}\left( {{\bf{x}},{\boldsymbol{\theta }}} \right)\\
		{\rm{s}}{\rm{.t}}{\rm{. }} & {f_i}\left( {{\bf{x}},{\boldsymbol{\theta }}} \right) \le 0,\\
		& i \in {\cal I},{\boldsymbol{\theta }} \in {\bf{\Theta }},
	\end{array}
	\label{equ:parametric-programming}
\end{equation}
where ${f_0}\left( {{\bf{x}},{\boldsymbol{\theta }}} \right)$ is a jointly convex objective function, ${f_i}\left( {{\bf{x}},{\boldsymbol{\theta }}} \right)$ are jointly convex constraints, $\mathcal{I}$ is the set of constraints, and $\bf{\Theta}$ is the feasible parameter space. According to the different forms of objective function and constraints such as linear, quadratic, and other non-linear, the map from parameter $\boldsymbol{\theta}$ to the optimal solution $\bf{x}\left(\boldsymbol{\theta}\right)$ and the value of objective function $V\left(\boldsymbol{\theta}\right)$ can be expressed as piecewise continuous rational polynomial fractions \cite{pappas2021exact}.
To solve the multi-parametric programming (\ref{equ:parametric-programming}), Lemma 2.1 in \cite{bemporad2006algorithm} is presented in the following.
\begin{lemma}
	Consider the multi-parametric programming (\ref{equ:parametric-programming}), if the objective and constraint functions are jointly convex in $({\bf x}, \boldsymbol{\theta})$ for all $i = {\cal I}$, then the feasible parameter set $\boldsymbol{\Theta}$ is convex, and the optimal value function $V(\boldsymbol{\theta})$ is a convex function of the parameter $\boldsymbol{\theta}$.
	\label{lemma:condition}
\end{lemma}
This property guarantees the regularity and continuity of the parametric solution, providing a theoretical foundation for constructing piecewise-affine representations of the optimizer.
The solution of (\ref{equ:parametric-programming}) can be expressed in an piecewise affine form of $\boldsymbol{ \theta}$ \cite{bemporad2006algorithm}, i.e.
\begin{equation}
{\bf x}\left( \boldsymbol{\theta}  \right) = \left\{ {\begin{array}{*{20}{c}}
	{{{\bf{X}}_1}{{\bf{M}}_1^{ - 1}}{{\left[ {1\;{\boldsymbol{\theta }}} \right]}^{\rm{T}}}},&{{\boldsymbol{\theta }} \in {{\boldsymbol{\Theta }}_1}},\\
	\vdots & \vdots \\
	{{{\bf{X}}_i}{{\bf{M}}_i^{ - 1}}{{\left[ {1\;{\boldsymbol{\theta }}} \right]}^{\rm{T}}}},&{{\boldsymbol{\theta }} \in {{\boldsymbol{\Theta }}_i}},\\
	\vdots & \vdots \\
	{{{\bf{X}}_{{k_\theta }}}{{\bf{M}}_{{k_\theta }}^{ - 1}}{{\left[ {1\;{\boldsymbol{\theta }}} \right]}^{\rm{T}}}},&{{\boldsymbol{\theta }} \in {{\bf{\Theta }}_{{k_\theta }}}},
	\end{array}} \right.
\end{equation}
where ${{\bf{\Theta }}_{{i }}} \subseteq {\bf \Theta}$ $\left(i=1,2,...,k_\theta\right)$ are given by the partitioning of the feasible parameter space $ {\bf{\Theta }}$ into polytopic regions, namely \emph{critical regions} ($CR$), ${\bf X}_i$ and ${\bf M}_i^{-1}$ are coefficient matrix for ${{\bf{\Theta }}_{{i }}}$, and $k_\theta$ is the number of critical regions. }

\begin{figure}
	\centering
	\includegraphics[width=0.6\linewidth]{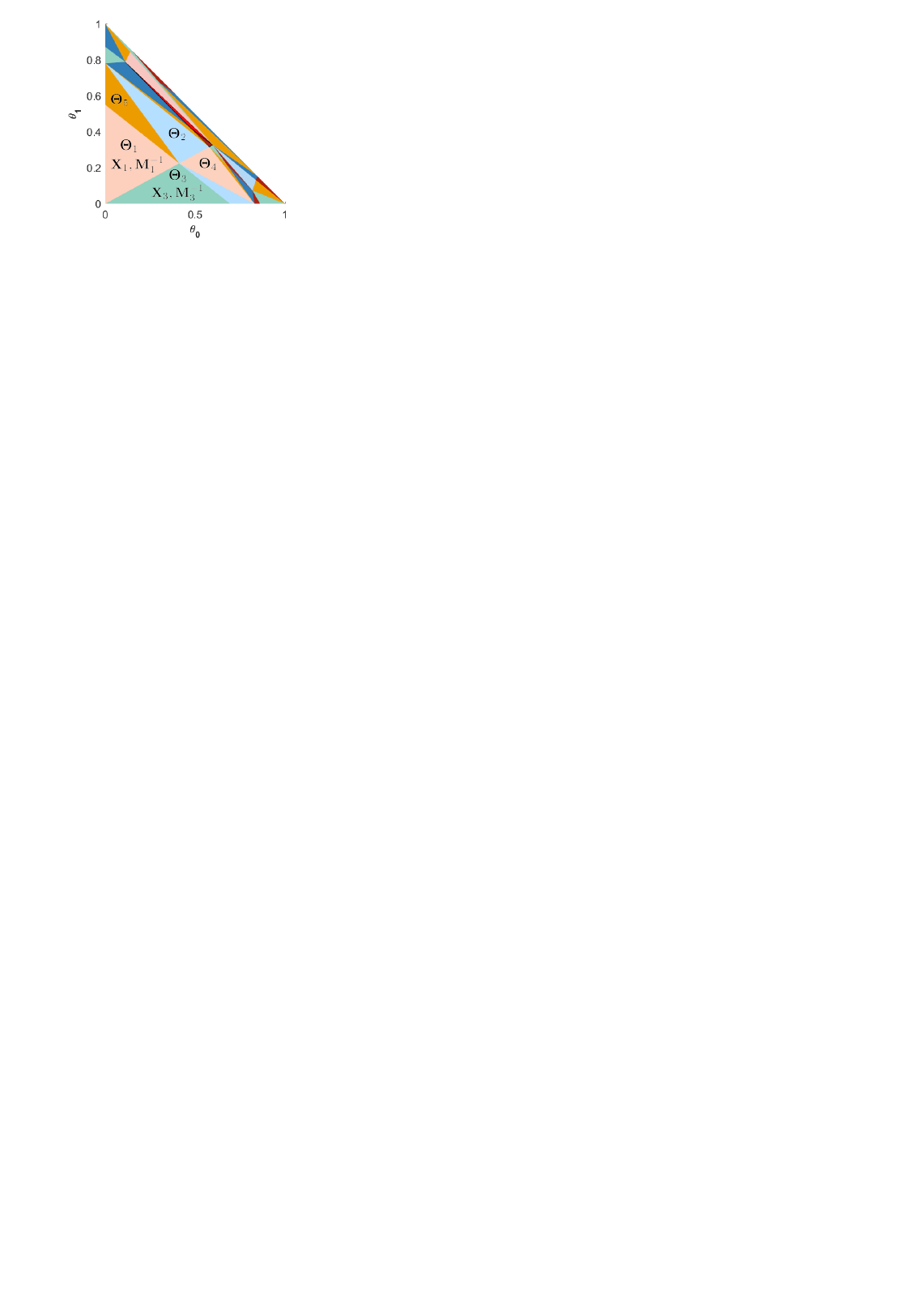}
	\caption{\textcolor{blue}{An example of critical regions with the parameter $\boldsymbol{\theta}=[\theta_0\;\theta_1] \in \boldsymbol{\Theta}$. Different colors represent different critical regions $\boldsymbol{ \Theta}_i (i=1,2,...,k_{\theta})$. Each critical region $\boldsymbol{ \Theta}_i$ has the corresponding parameters ${\bf X}_i$ and ${\bf M}_i^{-1}$. For the given $\boldsymbol{ \theta}\in \boldsymbol{\Theta}$, it will belong to a critical region $\boldsymbol{\Theta}_i$. Then, the optimal solution ${\bf x}(\boldsymbol{ \theta}) = {{{\bf{X}}_i}{{\bf{M}}_i^{ - 1}}{{\left[ {1\;{\boldsymbol{\theta }}} \right]}^{\rm{T}}}}$  is directly obtained without any optimization step.}}
	\label{fig:critical-region}
\end{figure}

Having obtained critical regions, as shown in Fig. \ref{fig:critical-region}, once the parameter $\boldsymbol{\theta}$ is set, the corresponding critical region is set, and the optimal solution ${\bf x}\left(\boldsymbol{\theta}\right)$ is obtained directly, which could efficiently reduce the computational complexity in real-time planning problem.

\subsection{Problem Formulation}
Suppose there is a swarm of robots with the number $N+1$ in an unknown obstacle environment, which is represented as a set \textcolor{blue}{ ${\cal N} = \left\{ i \right\}\left( i = 1,...,N,N+1\right).$} Let the start area and goal area be $\mathcal{C}_0$ and $\mathcal{C}_1$ in free space. 
The main problem is to plan an optimal virtual tube ${\cal T}^* = \left( {\mathcal{C}_0,\mathcal{C}_1,{\bf f},{{\bf h}^*}} \right)$ in free space of the obstacle-dense environment so that each robot $i\in \mathcal{N}$ is assigned an optimal trajectory ${\bf h}^*_i\left(t\right),t\in\left[ 0,T_i \right]$ in $\mathcal{T}^*$ which satisfies the following conditions:
\begin{itemize}
	\item The optimal trajectory ${\bf h}^*_i\left(t\right)$ is feasible for the dynamics of robot $i$, \textcolor{blue}{i.e., it satisfies the kinematic constraints, including the velocity bounds.}
	\item The optimal trajectory ${\bf h}^*_i(t)$ is defined as one that connects the start position ${\bf s}_i \in \mathcal{C}_0$ to the goal position ${\bf g}_i \in \mathcal{C}_1$, such that ${\bf h}^*_i(0) = {\bf s}_i$ and ${\bf h}^*_i(T_i) = {\bf g}_i$, \textcolor{blue}{while minimizing the total travel time $T_i$. In other words, this trajectory starts exactly at the start position ${\bf s}_i$, ends exactly at the goal position ${\bf g}_i$, and achieves the transfer in the shortest possible time.}
	
\item \textcolor{blue}{ Robot $i$ is stationary at both the start and the goal positions. Specifically, all time derivatives of the trajectory ${\bf h}^*_i(t)$ vanish at $t=0$ and $t=T_i$, that is, 
$
\frac{{{\rm d}^k}{\bf h}^*_i(0)}{{\rm d}t^k} = \frac{{{\rm d}^k}{\bf h}^*_i(T_i)}{{\rm d}t^k} = 0, \quad \forall k \ge 1,
$
which ensures that the velocity, acceleration, and higher-order derivatives are zero at both the start and goal, implying the robot remains at rest at these points.}
\end{itemize}
Meanwhile, a suitable controller is selected to make robot $i$ track its own trajectory and avoid other robots $j$ in the virtual tube, which is expressed as $\left\| {{{\bf{p}}_i}\left( t \right) - {{\bf{p}}_j}\left( t \right)} \right\| \ge 2{r_\text{s}},t \in \left[ {0,\textcolor{blue}{\max} \left( {{T_i},{T_j}} \right)} \right],\forall j \in {\cal N}\backslash i. $
\section{Architecture Overview}
\subsection{Framework}
\begin{figure*}
	\centering
	\includegraphics{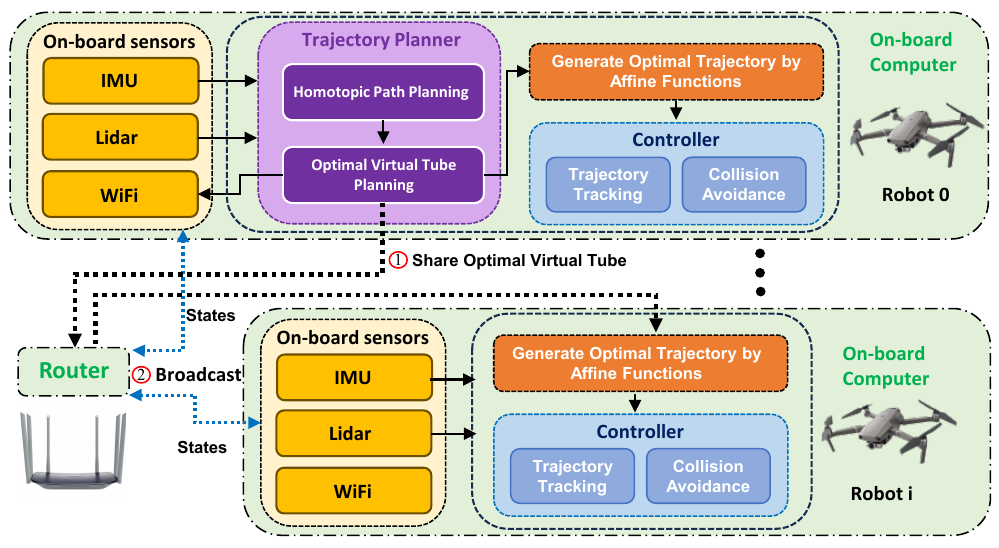}
	\caption{The framework of the proposed planning method. All robots have the same controller, and only Robot 1 has the planner block. The optimal virtual tube is planned by Robot 1 and shared with other robots by \textcircled{1}. The states of robots are shared by the broadcast \textcircled{2}. Other robots receive states and the optimal virtual tube to generate the optimal trajectory by affine functions. Then, the controllers of robots achieve trajectory tracking and collision avoidance.}
	\label{fig:framework}
\end{figure*}

The framework of the proposed planning method, as shown in Fig. \ref{fig:framework}, includes a planner and controller. \textcolor{blue}{Only the leader robot named Robot $1$ in the swarm has both planner block and controller block, and other followers named Robot $2 \sim N+1$ have only controller blocks.} The functions of all blocks are described in the following.

\textbf{Planner}: \textcolor{blue}{This module receives information from the on-board sensors and outputs the generated optimal virtual tube. Internally, it can be divided into two submodules: one for generating homotopic paths and another for generating the optimal virtual tube based on these paths. The final output is an optimal virtual tube, represented as a set of infinite homotopic trajectories, which serves as the reference trajectories for robot swarm.}

\textbf{Controller}: 
The robot receives the optimal virtual tube $\mathcal{T}^*$ via broadcast communication or internal interface \textcolor{blue}{from Robot $1$}. Then, the optimal trajectory ${\bf h}^*_i$ for robot $i$ is generated from optimal virtual tube $\mathcal{T}^*$ by affine functions. 
Subsequently, the controller combines trajectory tracking and collision avoidance to ensure that the robot can arrive at the goal position while avoiding other robots and tube boundaries.

The framework employs two communication methods:\textbf{ centralized optimal virtual tube planning} and \textbf{distributed control}. In centralized optimal virtual tube planning, the optimal virtual tube is planned on Robot 1 and then distributed to all robots in the swarm via the communication network denoted as \textcircled{1} in Fig. \ref{fig:framework}. In distributed control, each robot obtains the pose information of nearby robots and shares its own pose information denoted as \textcircled{2} in Fig. \ref{fig:framework} to achieve trajectory tracking and collision avoidance.

\subsection{Replanning Strategy}
\begin{figure}
	\centering
	\includegraphics{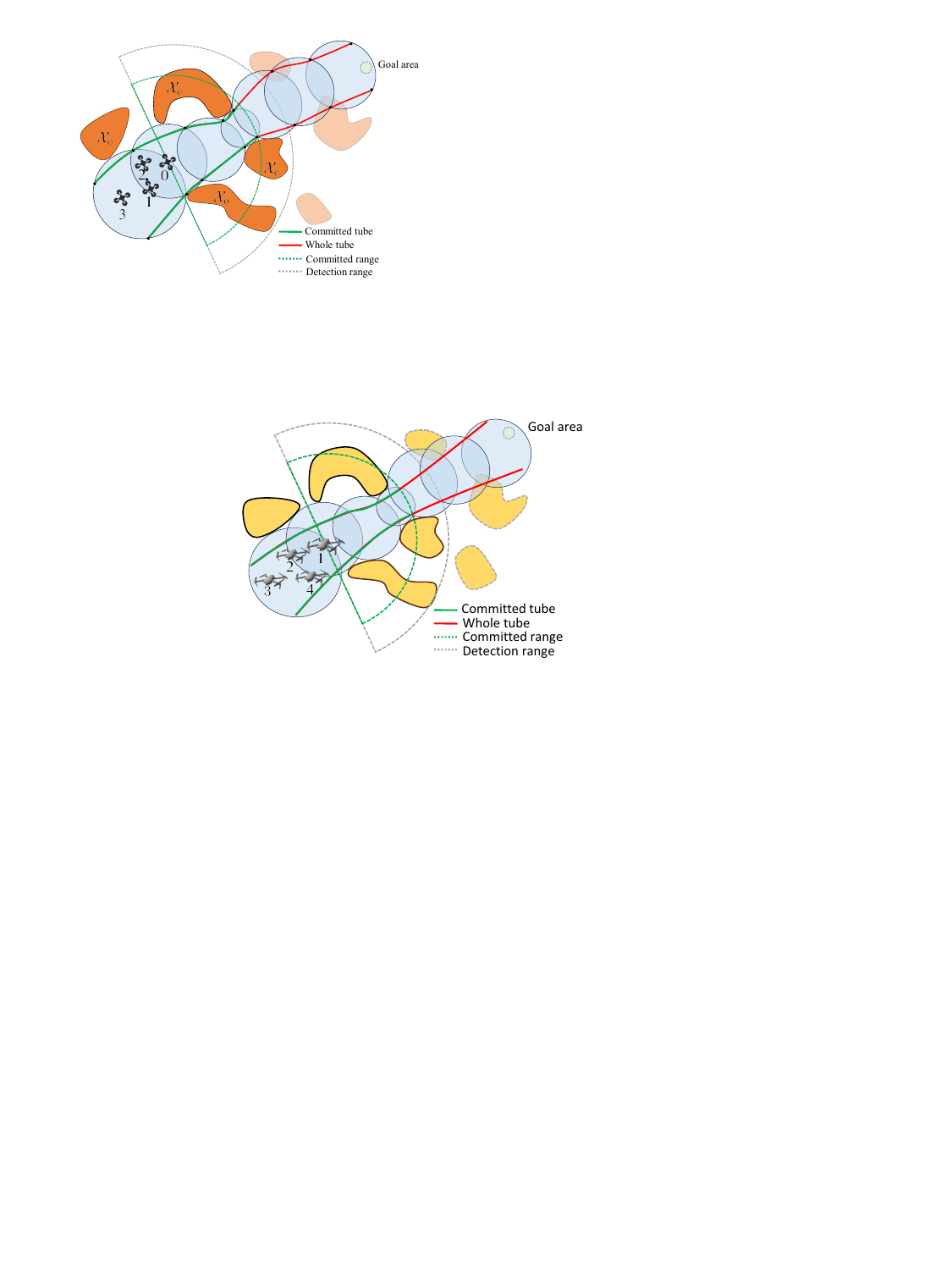}
	\caption{The replanning strategy of the proposed method. The known and unknown obstacles are colored orange and light orange, respectively. The boundary of the committed virtual tube and the whole virtual tube are represented by green and red curves. The sensing range and committed range are denoted by gray and green semicircles, respectively.}
	\label{fig:replanning-strategy}
\end{figure}

In the proposed method, since the optimal virtual tube is shared by all robots in the swarm, only a single shared virtual tube needs to be planned. As shown in Fig. \ref{fig:replanning-strategy}, all obstacles within the detection range are known, while it is assumed there are no obstacles outside the detection range. The planner uses the current obstacle information to plan a whole virtual tube to the goal area, as indicated by the red curves in Fig. \ref{fig:replanning-strategy}. However, the whole tube is likely to collide with unknown obstacles outside the detection range. To ensure safety, the portion of the tube within the detection range is designated as the committed tube, meaning it will no longer be updated. When robots in the swarm move beyond the committed tube, the virtual tube is replanned. Compared to other swarm trajectory update strategies, this method aligns the update frequency with that of single-robot trajectory planning \cite{gao2019flying,tordesillas2021faster}. Additionally, by delegating obstacle avoidance to the control level, the frequency of obstacle avoidance is significantly increased, enabling a faster response to obstacles.
\textcolor{blue}{\begin{rmk}
	The choice of leader robot depends on swarm size and sensing considerations: for small swarms, fixing a leader (whether at the front or centrally) is acceptable; for larger swarms, a dynamic leader-selection strategy is adopted in which the current leader monitors robots in the forward portion of the committed tube and computes the distance from the foremost robot to the leading boundary of tube — when this distance falls below a preset threshold the foremost robot is handed the leader role. This simple handover rule helps mitigate occlusion of the leader’s sensors and maintain forward perception.
\end{rmk}}
\section{Methods}
\textcolor{blue}{Optimal virtual tube planning comprises two stages: homotopic path planning and optimal virtual tube planning.} The first step, homotopic path planning, uses a suitable path planning algorithm to generate the homotopic paths for the virtual tube. Then, the second step, optimal virtual tube planning, plans the optimal virtual tube based on these path points. We detail these two steps in the following.
\subsection{Homotopic Path Planning}
An infinite homotopic path planning method, named Tube-RRT* \cite{mao2024tube}, has been proposed for virtual tube planning, which is inspired by the single-robot path planning methods in \cite{ren2022bubble,gao2019flying}. Tube-RRT* is a sample-based path planning method that samples intersected spheres in free space denoted by blue circles in Fig. \ref{fig:replanning-strategy} and plans a sequence of spheres from the start area to the goal area. The boundary paths $\boldsymbol{ \sigma}_k(k=1,2,...,k_c)$ for the virtual tube are selected in the sequence of spheres to be used for the trajectory optimization in the following.

\subsection{Optimal Virtual Tube Planning}
\textcolor{blue}{The module of optimal virtual tube planning uses the boundary paths $\{\boldsymbol{\sigma}_k\}$ as the input and generates the optimal virtual tube $\mathcal{T}^*$.}
In the previous work \cite{mao2023optimal}, the trajectory is parameterized by the piece-wise \textcolor{blue}{continuous} polynomials. There is a problem with this parameterization method, which is not theoretically able to guarantee that the trajectory remains within the safe region. Meanwhile, the desired total times for all trajectories in the virtual tube are supposed to be the same, which prevents further optimization of total time.
To overcome this problem, in this section, the trajectory in the optimal virtual tube is parameterized by piece-wise continuous B\'{e}zier curves. The convex hull property of B\'{e}zier curves can remain the trajectory within the convex hull of control points, which guarantees safety. 
\textcolor{blue}{Subsequently, a spatial-temporal trajectory optimization problem based on multi-parametric programming is proposed, where the core idea is to optimize key boundary trajectories to obtain an explicit characterization of all optimal trajectories, thereby improving both computational efficiency.}

We parameterize a trajectory, denoted as ${\bf{h}}\left(\left({\bf{q}}_0, {\bf{q}}_1\right), t\right)$, spanning from the start point ${\bf q}_0$ to the goal point ${\bf q}_1$, as a $p$-degree piece-wise continuous B\'{e}zier curve.  Specifically, ${\bf{h}}\left({\bf{q}}_0, {\bf{q}}_1, t\right)$ is a concatenation of a sequence of $M \in \mathbb{N}$ segments, ${\bf h}_1\left(t\right), {\bf h}_2\left(t\right), ...,{\bf h}_M\left(t\right)$. For segment $m\in\{1,2,...,M\}$, ${\bf h}_m\left(t\right)$ is characterized by control points ${{\bf{P}}_m} = [{{\bf{p}}_{m,0}}\quad{{\bf{p}}_{m,2}}\quad \cdots \quad{{\bf{p}}_{m,p}}] \in {\mathbb{R}^{3 \times \left( {p + 1} \right)}}$ and duration $\Delta {t_m} > 0$. Therefore, the whole trajectory ${\bf{h}}\left(\left({\bf{q}}_0, {\bf{q}}_1\right), t\right)$ could be expressed as
\begin{equation}
	{\bf{h}}\left( {\left( {{{\bf{q}}_0},{{\bf{q}}_1}} \right),t} \right)  \buildrel \Delta \over =  {{\bf{h}}_m}\left( {t - \sum\limits_{j = 1}^{m - 1} {\Delta {t_j}} } \right),
\end{equation}
where $t \in \left[ {\sum\nolimits_{j = 1}^{m - 1} {\Delta {t_j}} ,\sum\nolimits_{j = 1}^m {\Delta {t_j}} } \right]$, ${\bf q}_0 = {\bf p}_{1,0}$. ${\bf q}_1={\bf p}_{M,p}$,  
\[{{\bf{h}}_m}\left( {t - \sum\limits_{j = 1}^{m - 1} {\Delta {t_j}} } \right) = {{\bf{C}}_m}\boldsymbol{\beta}^{\rm T} \left( t \right),\] $\boldsymbol{\beta} \left( t \right) = {\left[ {1,t,{t^2},...,{t^{{p}}}} \right]}, t \in \left[0,\Delta {t_m}\right]$, ${{\bf{C}}^{\rm T}_m} = {{\bf{S}}_{\Delta {t_m}}}{\bf{M}}{{\bf{P}}_m}^{\rm{T}}$. 

The trajectory ${\bf{h}}\left(\left({\bf{q}}_0, {\bf{q}}_1\right), t\right)$, or called ${\bf h}\left(t\right)$ for simply, is constrained in a sequence of intersected spheres given by Section V-A. The terminals for every segment of the trajectory are placed in the intersection areas of adjoint spheres. This ensures that each segment is confined within distinct spheres, as illustrated in Fig. \ref{fig:path-point-selection-strategy}. Consequently, the avoidance constraints can be expressed as:
\begin{equation}
	{S_m}\left( {{{\bf{h}}_m}\left( t \right)} \right) \le 0, m\in 1,...,M.
\end{equation}
Together with dynamic feasibility in the form of the derivatives of the trajectory, the spatial-temporal trajectory optimization is formulated by
\begin{subequations}
	\begin{align}
		\mathop {\min }\limits_{{\bf{P}},{\bf{t}}}& {\rm{ }}\int_0^{\sum\nolimits_{m = 1}^M {\Delta {t_m}} } {{{\left\| {{{\bf{h}}^{\left( d \right)}}\left( t \right)} \right\|}^2}{\rm{d}}t}  + {\rho _t}\sum\nolimits_{m = 1}^M {\Delta {t_m}} \\
		{\rm{s}}{\rm{.t}}{\rm{. }}\,&{{\bf{h}}^{\left[ {d - 1} \right]}}\left( 0 \right) = {{{\bf{\bar p}}}_{1,0}},{{\bf{h}}^{\left[ {d - 1} \right]}}\left( {\sum\nolimits_{m = 1}^M {\Delta {t_m}} } \right) = {{{\bf{\bar p}}}_{M,p}},\\
		&{\bf{h}}_m^{\left[ {d - 1} \right]}\left( {\Delta {t_m}} \right) = {\bf{h}}_{m + 1}^{\left[ {d - 1} \right]}\left( 0 \right),\\
		& - {{\bf{i}}_m} \le {\bf{h}}_m^{\left[ {d - 1} \right]}\left( t \right) \le {{\bf{i}}_m},\\
		& {S_m}\left( {{{\bf{h}}_m}\left( t \right)} \right) \le 0,\\
		& m=1,2,...,M,\, t\in \left[0,\Delta t_m\right],
		\end{align}
		\label{equ:spatial-temporal-optimization}
\end{subequations}
\unskip where $\rho_t>0$ is the weight coefficient, ${\bf{h}}_m^{\left[ {d - 1} \right]}\left( t \right) = {[ {{{\bf{h}}_m}{{\left( t \right)}^{\rm{T}}},{\bf{h}}_m^{\left( 1 \right)}{{\left( t \right)}^{\rm{T}}},...,{\bf{h}}_m^{\left( {d - 1} \right)}{{\left( t \right)}^{\rm{T}}}} ]^{\rm{T}}}$, 
${{{\bf{\bar p}}}_{1,0}}$, ${{{\bf{\bar p}}}_{M,p}}$ are start and goal states up to $\left(d-1\right)$-order derivatives, ${\bf i}_m$ is the dynamic bounds.

\textcolor{blue}{All trajectories within the optimal virtual tube are generated by parameters $({\bf P}^*,{\bf t}^*)$ that are the optimal solutions of problem (\ref{equ:spatial-temporal-optimization}).}
However, directly solving this problem is highly challenging, and solving infinite optimization problems involving an infinite number of trajectories within the optimal virtual tube is infeasible.
To streamline the programming associated with (\ref{equ:spatial-temporal-optimization}), it is decomposed into two distinct steps.
In the first step, path points are generated through spatial trajectory optimization, ensuring trajectories remain within the virtual tube and minimizing control efforts, capitalizing on the convex hull property of the B\'{e}zier curve. In the second step, feasible dynamics of robots are considered in the temporal trajectory optimization. Subsequently, we employ multi-parametric programming to obtain an infinite set of optimal trajectories by solving a finite number of programmings.
\subsubsection{Spatial Trajectory Optimization}
\begin{figure}
 	\centering
 	\includegraphics[width=\linewidth]{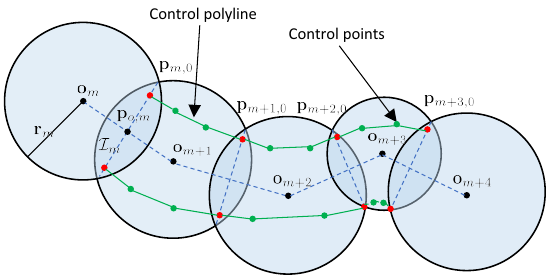}
 	\caption{Examples of spatial trajectory optimization (\textcolor{blue}{adapted and modified from \cite{mao2024tube}}). The red points and light blue points represent control points. Red points, especially, are terminals of segments that are fixed in the intersection areas of adjoint spheres. The light blue lines are the control polylines.}
 	\label{fig:path-point-selection-strategy}
 \end{figure}

The boundary trajectories of the virtual tube are generated first. Before formulating the spatial trajectory optimization, a suitable terminal path points selection strategy is designed in this section to make the trajectory in the spheres. As shown in Fig. \ref{fig:path-point-selection-strategy}, terminal path points ${\bf p}_{m,0},{\bf p}_{m,p}$ for any segment $m$ of the trajectory are constrained within the intersection areas of adjoint spheres to maintain trajectory in free space.
Suppose there is a sequence of intersected spheres centered in ${\bf o}_m$ with radius $r_m$ $\left(m=1,...,M\right)$. 
The intersection area of adjoint spheres centered in ${\bf o}_m$ and ${\bf o}_{m+1}$ respectively is denoted by ${{\cal I}_{m}}$.
Let $^k{\bf p}_{1,0} \in {\mathcal{I}_1}$ $\left(k\in1,2,...,k_c\right)$ be the start point, where $k_c$ is the number of boundary trajectories and is designed manually. Then, other start points in $\mathcal{I}_m$ are expressed as ${^k{\bf{p}}_{m,0}} \in {{\cal I}_m}.$ 

To initialize the time allocation $\left\{ {\Delta {t_m}} \right\}$ by the chord length parameterization method \cite{piegl2000surface}, the initial positions of terminal points are needed. For convenience, the initial terminal points are set in the maximum intersection plane ${{\cal I}_{m,\text{max}}}$ in the intersection area of adjoint spheres ${{\cal I}_{m}}$. The ${{\cal I}_{m,\text{max}}}$ is expressed as
\begin{equation}
	{\mathcal{I}_{m,\text{max} }}\left(\rho,\varphi \right) = {{\bf{p}}_{o,m}} + \rho{\lambda _{m}}\left( {{{\bf{n}}_{o,m}}\cos \varphi  + {{\bf{b}}_{o,m}}\sin \varphi } \right),
\end{equation}
where $\varphi \in \left[ {0,2\pi } \right]$, $\rho \in \left[ 0, 1 \right] $, ${{\bf{v}}_{o,m }}, {{{\bf{n}}_{o,m}}}$ and ${{{\bf{b}}_{o,m}}}$ are the unit tangent vector, unit normal vector and unit abnormal vector of the line $\overline {{{\bf{o}}_{m + 1}}{{\bf{o}}_m}} $ respectively,
\[{{\bf{p}}_{o,m}} = {{\bf{o}}_{m}} + \frac{{{r^2_{m + 1}} - {r^2_m} + {{\left\| {{{\bf{o}}_{m + 1}} - {{\bf{o}}_m}} \right\|}^2}}}{{2\left\| {{{\bf{o}}_{m + 1}} - {{\bf{o}}_m}} \right\|}}{{\bf{v}}_{o,m}},\]
\[{\lambda _{m}} = \sqrt {{r^2_{m + 1}} - \frac{{{r^2_{m + 1}} - {r^2_m} + {{\left\| {{{\bf{o}}_{m + 1}} - {{\bf{o}}_m}} \right\|}^2}}}{{2\left\| {{{\bf{o}}_{m + 1}} - {{\bf{o}}_m}} \right\|}}}. \]
After that, the time allocations $\left\{ {\Delta {t_m}} \right\}$ for all boundary trajectories are set as the same by the arithmetic mean \cite{STROTHOTTE2002203}.

\begin{figure}
	\centering
	\includegraphics{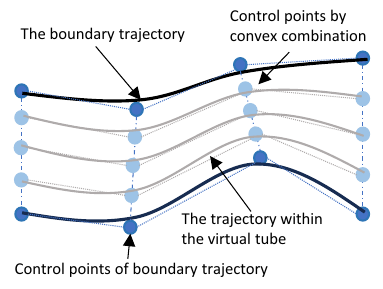}
	\caption{A method to generate the trajectories within the virtual tube. The dark blue circles and light blue circles represent the control points of boundary trajectories and control points of trajectories within the virtual tube, respectively. The black curves are boundary trajectories, and the gray curves are trajectories within the virtual tube.}
	\label{fig:linear-interpolation-control-points}
\end{figure}

Subsequently, the spatial trajectory optimization for a trajectory ${\bf h}\left(t\right)$ in the virtual tube is formulated by
\begin{subequations}
	\begin{align}
		\mathop {\min }\limits_{{\bf{P}}}& {\rm{ }}\int_0^{\sum\nolimits_{m = 1}^M {\Delta {t_m}} } {{{\left\| {{{\bf{h}}^{\left( d \right)}}\left( t \right)} \right\|}^2}{\rm{d}}t}\\
		{\rm{s}}{\rm{.t }}
		&{\rm{. }}{{\bf{h}}^{\left[ {d - 1} \right]}}\left( 0 \right) = {{{\bf{\bar p}}}_{1,0}},{{\bf{h}}^{\left[ {d - 1} \right]}}\left( {\sum\nolimits_{m = 1}^M {\Delta {t_m}} } \right) = {{{\bf{\bar p}}}_{M,p}}, \\
		& {\bf{h}}_m^{\left[ {d - 1} \right]}\left( {\Delta {t_m}} \right) = {\bf{h}}_{m + 1}^{\left[ {d - 1} \right]}\left( 0 \right),\\
		& {S_m}\left( {{{\bf{h}}_m}\left( t \right)} \right) \le 0,\\
		& - {{\bf{i}}_\text{m}} \le {\bf{h}}_i^{\left[ {d - 1} \right]}\left( t \right) \le {{\bf{i}}_\text{m}},\\
		& m=1,2,...,M,\, t\in \left[0,\Delta t_m\right].
	\end{align}
	\label{equ:spatial-trajectory-optimization}
\end{subequations}
Thus, the $k_c$ boundary trajectories can be generated by (\ref{equ:spatial-trajectory-optimization}). And for any trajectory ${\bf h}_{\boldsymbol{\eta}} \left(t\right)$ in the virtual tube whose start point 
is 
\[{{\bf{p}}_{1,0}}\left( \boldsymbol{\eta}  \right) = \sum\nolimits_{k = 1}^{{k_c}} {{\eta _k}^k{{\bf{p}}_{1,0}}} ,\sum\nolimits_{k = 1}^{{k_c}} {{\eta _k}}  = 1,{\eta _k} \ge 0,\]
according to \emph{Lemma 2} in \cite{mao2023optimal},
the control points of any segment $m$ of trajectory ${\bf h}_{\boldsymbol{\eta}} \left(t\right)$ generated by (\ref{equ:spatial-trajectory-optimization}) are expressed as
\begin{equation}
	{{\bf{P}}_m}\left( \boldsymbol{\eta}  \right) = \sum\nolimits_{k = 1}^{{k_c}} {{\eta _k}^k{{\bf{P}}_m}} ,\sum\nolimits_{k = 1}^{{k_c}} {{\eta _k}}  = 1,{\eta _k} \ge 0.
	\label{equ:control-points-for-any-trajectory}
\end{equation}
According to (\ref{equ:control-points-for-any-trajectory}), control points for any trajectory in the virtual tube are expressed as the convex combination of control points of boundary trajectories, as shown in Fig. \ref{fig:linear-interpolation-control-points}.
\subsubsection{Temporal Trajectory Optimization}
The control points of the boundary trajectories are generated by the above subsection. 
The object of the temporal trajectory optimization problem is to arrive at a goal as fast as possible, in other words, to minimize the total time. Thus, the optimization problem is expressed as
\begin{subequations}
	\begin{align}
		 \mathop {\min }\limits_{{\bf{t}}} &{\rm{ }}\sum\nolimits_{m= 1}^M {\Delta {t_m}} \\
		{\rm{s}}{\rm{.t}}{\rm{. }} & - {{\bf{i}}_\text{m}} \le {\bf{h}}_m^{\left[ {d - 1} \right]}\left( t \right) \le {{\bf{i}}_\text{m}},\\
		& {\bf{h}}_m^{\left[ {d - 1} \right]}\left( {\Delta {t_m}} \right) = {\bf{h}}_{m + 1}^{\left[ {d - 1} \right]}\left( 0 \right),\\
		& m=1,2,...,M,\, t\in \left[0,\Delta t_m\right].
	\end{align}
	\label{equ:temp-prob}
\end{subequations}
\unskip In this problem, because of the mass point model of the robot in (\ref{equ:robotmodel}), we only constraint the velocity so that the problem (\ref{equ:temp-prob}) is transformed into a linear programming:
\begin{subequations}
	\begin{align}
		\mathop {\min }\limits_{{\bf{t}}} &{\rm{ }}\sum\nolimits_{m = 1}^M {\Delta {t_m}} \\
		{\rm{s}}{\rm{.t}}{\rm{. }} & - {{\bf{v}}_\text{m}} \le \frac{p}{{\Delta {t_m}}}\left( {{{\bf{p}}_{m,k + 1}} - {{\bf{p}}_{m,k}}} \right) \le {{\bf{v}}_\text{m}},\\
		& \frac{{\left( {{{\bf{p}}_{m,p}} - {{\bf{p}}_{m,p - 1}}} \right)}}{{\Delta {t_m}}} = \frac{{\left( {{{\bf{p}}_{m + 1,1}} - {{\bf{p}}_{m + 1,0}}} \right)}}{{\Delta {t_{m + 1}}}},\\
		& \Delta {t_m} > 0,m = 1,2,...,M,k=0,...,p-1,
	\end{align}
	\label{equ:traj_plan_prblm}
\end{subequations}
\unskip where ${\bf v}_\text{m}$ are the velocity limitations of the robot. Thus, (\ref{equ:traj_plan_prblm}) could be expressed as the standard form, which is expressed as
\textcolor{blue}{\begin{equation}
\begin{array}{rl}
\mathop {\min }\limits_{\bf{x}}& {{\bf{c}}^{\rm{T}}}{\bf{x}}\\
{\rm{s}}.{\rm{t}}.&{\rm{ }}{{\bf{A}}_1}{\bf{x}} \le {{\bf{b}}_1},\\
&{\rm{ }}{{\bf{A}}_1}{\bf{x}} \le -{{\bf{b}}_1},\\
&{{\bf{A}}_3}{\bf{x}} = {\bf{0}},
\end{array}
\end{equation}
where ${\bf{x}} = {\left[ {
			{\Delta {t_1}}\ {\Delta {t_2}}\  \cdots \ {\Delta {t_M}}
	} \right]^{\rm{T}}}$, 
\[{{\bf{A}}_1} = \left[ {\begin{array}{*{20}{c}}
	{{{\bf{A}}_{1,1}}}\\
	\vdots \\
	{{{\bf{A}}_{1,m}}}\\
	\vdots \\
	{{{\bf{A}}_{1,M}}}
	\end{array}} \right],{{\bf{b}}_1} = \left[ {\begin{array}{*{20}{c}}
	{{{\bf{b}}_{1,1}}}\\
	\vdots \\
	{{{\bf{b}}_{1,m}}}\\
	\vdots \\
	{{{\bf{b}}_{1,M}}}
	\end{array}} \right],\]
\[{{\bf{A}}_{1,m}} = \left[ {\begin{array}{*{20}{c}}
	{\bf{0}}& \cdots &{{{\bf{v}}_m}}& \cdots &{\bf{0}}\\
	\vdots & \vdots & \vdots & \vdots & \vdots \\
	{\bf{0}}& \cdots &{{{\bf{v}}_m}}& \cdots &{\bf{0}}
	\end{array}} \right],\]
\[{{\bf{b}}_{1,m}} =  - p\left[ {\begin{array}{*{20}{c}}
	{{{\bf{p}}_{m,1}} - {{\bf{p}}_{m,0}}}\\
	\vdots \\
	{{{\bf{p}}_{m,p - 1}} - {{\bf{p}}_{m,p - 2}}}
	\end{array}} \right].\]
}   
Because there are $k_c$ boundary trajectories in the virtual tube. The optimization problem of the $k$-th boundary trajectory ${\bf h}_k\left(t\right)$ is expressed as 
\textcolor{blue}{\begin{equation}
\begin{array}{rl}
\mathop {\min }\limits_{\bf{x}}& {{\bf{c}}^{\rm{T}}}{\bf{x}}\\
{\rm{s}}.{\rm{t}}.&{\rm{ }}{{\bf{A}}_{1,k}}{\bf{x}} \le {{\bf{b}}_{1,k}},\\
&{{\bf{A}}_{2,k}}{\bf{x}} = {\bf{0}}.
\end{array}
\end{equation}}
For any trajectory ${\bf h}_{\boldsymbol{\theta}}\left(t\right)$ within the virtual tube, the control points are the convex combination of the control points of the boundary trajectories, namely ${{\bf{P}}_{\boldsymbol{\theta}} } \in {\rm{conv}}\left( {\left\{ {{^k{\bf{P}}}} \right\}} \right)$. Thus, the optimization problem of the trajectory ${\bf h}_{\boldsymbol{\theta}}\left(t\right)$ is expressed as
\textcolor{blue}{\begin{subequations}
\begin{align}
\mathop {\min }\limits_{\bf{x}}\; & {{\bf{c}}^{\rm{T}}}{\bf{x}}\\
{\rm{s}}.{\rm{t}}. \; & {{\bf{A}}_1}{\bf{x}} \le {{\bf{b}}_1} + {\bf{F \boldsymbol{\theta} }}, \label{equ:virtual_tube_generation_b}\\
& {{\bf{A}}_2}\left( \boldsymbol{\theta}  \right){\bf{x}} = {\bf{0}}, \label{equ:virtual_tube_generation_c}
\end{align}
\label{equ:virtual_tube_generation}
\end{subequations}
where ${\bf{F}} = \left[ {{{\bf{b}}_{1,2}} - {{\bf{b}}_{1,1}}\; \cdots \;{{\bf{b}}_{1,{k_c}}} - {{\bf{b}}_{1,1}}} \right]$,
${{\bf{A}}_2}\left( {\boldsymbol{\theta }} \right) = {{\bf{A}}_{2,1}} + \sum\nolimits_{k = 1}^{{k_c} - 1} {{\theta _k}{{\bf{A}}_{2,k + 1}}}$, $\sum\nolimits_{k = 1}^{{k_c-1}} {{\theta _k} = 1}$, ${\theta _k} \ge 0$.} The optimal value of (\ref{equ:virtual_tube_generation}) is denoted as $V^*\left(\boldsymbol{\theta}\right)$. 
\textcolor{blue}{Since the parameter domain $\boldsymbol{\theta}$ is continuous and bounded, directly computing the optimal value for every possible $\boldsymbol{\theta}$ is infeasible. Therefore, the multi-parametric programming described in Section III-D is required to iteratively partition the parameter domain and derive explicit parametric expressions of the optimal solutions.} 
A recursive algorithm, detailed in \emph{Algorithm \ref{alg:paramprog}}, for approximate multi-parametric programming \cite{bemporad2006algorithm} could be used to find the critical regions $CR_\theta$ for all ${\boldsymbol{\theta}}$ and the corresponding explicit optimal solutions ${\bf{x}}_{C{R_\theta }}^*\left( {\boldsymbol{\theta }} \right) = {\bf{X}}{{\bf{M}}^{ - 1}}{\left[ {
		1\ \boldsymbol{\theta} 
	} \right]^{\rm{T}}}$ within the error bound $\epsilon$, where ${\bf X}$ and $ {\bf M}$ are related to the corresponding region in $CR_{\theta}$. \textcolor{blue}{The steps of \emph{Algorithm \ref{alg:paramprog}} are detailed in the following.} 

\textcolor{blue}{\emph{Step 1} (lines 1–2): Compute the optimal values ${\bf v}_{\text{opt}} = [V^*(\boldsymbol{\theta}_1),V^*(\boldsymbol{\theta}_2),...,V^*(\boldsymbol{\theta}_{k_c})]$ and the corresponding optimal solutions ${\bf x}^*_1, {\bf x}^*_2,...,{\bf x}^*_{k_c}$ of (\ref{equ:virtual_tube_generation}) for the $k_c$ boundary trajectories. Each parameter vector in $\boldsymbol{\theta}_1,\boldsymbol{\theta}_2,...,\boldsymbol{\theta}_{k_c} \in CR_\theta$ defines a specific instance of (\ref{equ:virtual_tube_generation}) associated with one boundary trajectory. For example, the parameter vector ${\boldsymbol{ \theta}_1} = [0,0,...,0]^{\text{T}}\in \mathbb{R}^{k_c-1}$ can be substituted into (\ref{equ:virtual_tube_generation}) to obtain the optimization problem for the first boundary trajectory. Using these results, construct two matrices, ${\bf M}$ and ${\bf X}$, which will be used in the subsequent steps.
}

\textcolor{blue}{\emph{Step 2} (line 3): Check whether the matrix $\bf M$ is singular. If it is singular, this indicates that the selected parameter vectors cannot form a simplex; that is, at least one parameter vector cannot serve as a vertex of the simplex. If the matrix $\bf M$ is non-singular, then the selected parameter vectors can serve as vertices to form a new partition (critical region $\boldsymbol{ \Theta}$) in the parameter domain, and the algorithm proceeds to the next step.
}

\textcolor{blue}{\emph{Step 3} (line 4): Within the critical region $\boldsymbol{ \Theta}$ obtained in the previous step, each $\boldsymbol{ \theta} \in \boldsymbol{\Theta}$ corresponds to an optimization problem (\ref{equ:virtual_tube_generation}) for one trajectory and thus has a true optimal value $V^*(\boldsymbol{ \theta})$. The goal of this algorithm is to approximate $V^*(\boldsymbol{ \theta})$ using a convex combination of the optimal values at the vertices of the critical region $\boldsymbol{ \Theta}$, which can be expressed as
\begin{equation}
\tilde{V}(\boldsymbol{ \theta}) = {{\bf{v}}_{{\rm{opt}}}}{{\bf{M}}^{ - 1}}\left[ {\begin{array}{*{20}{c}}
	1\\
	\boldsymbol{\theta}
	\end{array}} \right].
\end{equation}
And, the error between $V^*(\boldsymbol{ \theta})$ and $\tilde{V}(\boldsymbol{ \theta})$ is expressed as $\tilde{\epsilon} = \tilde{V}(\boldsymbol{ \theta}) - V^*(\boldsymbol{ \theta}) = \tilde{V}(\boldsymbol{ \theta}) - {{\bf{c}}^{\rm{T}}}{\bf{x}}$. To properly control the approximation error $\tilde{\epsilon}$, an error bound $\epsilon$ must be specified, and all errors within this critical region $\boldsymbol{ \Theta}$ must be kept below this bound. Therefore, it is necessary to ensure that the maximum approximation error within the current critical region is still below this threshold. The task of finding the maximum error inside the region can be achieved by formulating the following optimization problem:	
\begin{subequations}
\begin{align}
{\bar \epsilon} = \mathop {\max }\limits_{{\bf{x}},\boldsymbol{\theta} }  & \; \tilde{V}(\boldsymbol{ \theta})- {{\bf{c}}^{\rm{T}}}{\bf{x}}\\
{\rm{s}}{\rm{.t}}{\rm{.  }}&{{\bf{A}}_1}{\bf{x}} \le {{\bf{b}}_1} + {\bf{F \boldsymbol{\theta} }},\\
& {{\bf{A}}_2}\left( \boldsymbol{\theta}  \right){\bf{x}} = {\bf{0}},\\ \label{equ:step3_c}
&{{\bf{M}}^{ - 1}}\left[ {\begin{array}{*{20}{c}}
	1\\
	\boldsymbol{\theta} 
	\end{array}} \right]  \ge 0.	
\end{align}
\label{equ:step3}
\end{subequations}}
\textcolor{blue}{However, the constraints (\ref{equ:step3_c}) are bilinear in both ${\bf x}$ and $\boldsymbol{\theta}$, which violates the joint convexity requirement in Lemma \ref{lemma:condition}. To address this issue, the constraints must be reformulated into a jointly convex representation. By applying an SDP relaxation \cite{lee2016sequential} to the bilinear constraints, joint convexity can be achieved, and the constraints are transformed into convex ones. Specifically, an augmented vector is introduced as ${\bf{z}} = {[
		{{\boldsymbol{ \theta} ^{\rm{T}}}},{{{\bf{x}}^{\rm{T}}}}
		]^{\rm{T}}}$, and the lifted matrix is defined as
\begin{equation}
{\bf{Z}} = {\bf{z}}{{\bf{z}}^{\rm{T}}},{\bf{Z}}\succeq 0.
\end{equation}
Under this lifting, every bilinear term such as $\theta_kx_i$ becomes a linear entry of the lifted matrix $\bf Z$. Consequently, the nonlinear equality constraint (\ref{equ:virtual_tube_generation_c}) can be expressed as a set of quadratic forms in $\bf z$:
\begin{equation}
{{\bf{z}}^{\rm{T}}}{{\bf{Q}}_m}{\bf{z}} = 0,
\end{equation}
where each matrix ${\bf{Q}}_m$ encodes the coefficients of the corresponding row of (\ref{equ:step3_c}). After lifting, these constraints become linear equalities 
\begin{equation}
{\rm{Tr}}\left( {{{\bf{Q}}_m}{\bf{Z}}} \right) = {\bf 0}.
\end{equation}
The remaining constraints in (\ref{equ:step3}) also become linear in the lifted variable $\bf Z$ because both $\bf x$ and $\boldsymbol{ \theta}$ correspond to specific entries of the vector $\bf z$. Therefore, the original optimization problem (\ref{equ:step3}) is relaxed into the following SDP:
\begin{equation}
\begin{array}{rl}
{\bar \epsilon} = \mathop {\max }\limits_{\bf Z} &\;\tilde V({\boldsymbol{\theta }}) - {{\bf{c}}^{\rm{T}}}{{\bf{Z}}_x}\\
{\rm{s}}.{\rm{t}}.&{{\bf{A}}_1}{{\bf{Z}}_x} \le {{\bf{b}}_1} + {\bf{F}}{{\bf{Z}}_\theta },\\
&{\rm{Tr}}\left( {{{\bf{Q}}_m}{\bf{Z}}} \right) = 0,\forall m,\\
&{{\bf{M}}^{ - 1}}\left[ {\begin{array}{*{20}{l}}
	1\\
	{{{\bf{Z}}_\theta }}
	\end{array}} \right] \ge 0,\\
&{\bf{Z}} \succeq 0.
\end{array}
\label{equ:SDP}
\end{equation}
where ${\bf Z}_x$ and ${\bf Z}_\theta$ denote the blocks of ${\bf Z}$ corresponding to the components of $\bf x$ and $\boldsymbol{ \theta}$, respectively. By solving this optimization problem (\ref{equ:SDP}), the maximum approximation error $\bar \epsilon$ and its corresponding optimizer $({{\bf{\bar x}}},{\boldsymbol{\bar \theta}})$ can be obtained.
}

\textcolor{blue}{\emph{Step 4} (lines 5-12): Determine whether the error $\bar \epsilon$ is below the error bound $\epsilon$.
If it is below the $\epsilon$, the approximation $\tilde{V}({\boldsymbol{ \theta}})$ over the current partition $\boldsymbol{ \Theta}$ is acceptable, and the corresponding parameters $\bf X$ and $\bf M$ can be directly returned.
If it exceeds the $\epsilon$, this indicates that the current partition $\boldsymbol{\Theta}$ requires further refinement by adding the parameter $\bar {\boldsymbol{\theta}}$ corresponding to $\bar\epsilon$ as a new vertex, after which the algorithm is iteratively invoked again.}
\begin{algorithm}
	\caption{An approximate multi-parametric programming}
	\begin{algorithmic}[1]
		\REQUIRE (i) parameter vectors $\boldsymbol{\theta}_1,\boldsymbol{\theta}_2,...,\boldsymbol{\theta}_{k_c} \in CR_\theta$; (ii) the optimal values ${\bf v}_{\text{opt}} = [V^*(\boldsymbol{\theta}_1),V^*(\boldsymbol{\theta}_2),...,V^*(\boldsymbol{\theta}_{k_c})]$ of (\ref{equ:virtual_tube_generation}) corresponding to the parameter vectors; (iii) vector ${\bf x}^*_1, {\bf x}^*_2,...,{\bf x}^*_{k_c}$ such that ${\bf x}^*_k$ is the optimal solution of (\ref{equ:virtual_tube_generation}) for all $k=1,2,...,k_c$; (iv) the error bound $\epsilon$.
		\ENSURE  the parameters ${\bf M}$ and ${\bf X}$ corresponding to the critical regions $CR_\theta$.
		\STATE Build ${\bf M} = \left[ {\begin{array}{*{20}{c}}
				1&1& \cdots &1\\
				{{\boldsymbol{\theta} _1}}&{{\boldsymbol{\theta} _2}}& \cdots &{{\boldsymbol{\theta} _{{k_c}}}}
		\end{array}} \right]$;
		\STATE Build ${\bf X} = \left[ {\begin{array}{*{20}{c}}
				{{{\bf{x}}^*_1}}&{{{\bf{x}}^*_2}}& \cdots &{{{\bf{x}}^*_{{k_c}}}}
		\end{array}} \right]$;
		\IF{$\bf M$ is nonsingular}
		\STATE Compute the optimum ${\tilde \epsilon }$ and an optimizer $({{\bf{\bar x}}},{\boldsymbol{\bar \theta}})$ of (\ref{equ:SDP});
		\IF{$\tilde\epsilon > \epsilon$}
		\FOR{$k=1,2,...,k_c-1$}
		\STATE replace ${\boldsymbol{\theta}_k}$ by $\boldsymbol{\bar\theta}$, $V^*(\boldsymbol{\theta}_k)$ by $V(\boldsymbol{\bar \theta})$, and ${\bf x}^*_k$ by $\bar{\bf x}$;
		\STATE call this algorithm;
		\ENDFOR
		\ELSE
		\RETURN $\left( {{\bf{M}},{\bf{X}}} \right)$;
		\ENDIF
		\ENDIF
	\end{algorithmic}
	\label{alg:paramprog}
\end{algorithm}
\textcolor{blue}{Through the recursive calls of the \emph{Algorithm \ref{alg:paramprog}}, critical regions $CR_\theta$ that satisfies the error bound $\epsilon$ and its corresponding parameters are eventually obtained, thereby establishing the mapping between $\boldsymbol{ \theta}$ and the optimal solution $\bf x$ (which represents the time interval $\bf t$ in temporal trajectory optimization).}
Combining spatial and temporal trajectory optimizations, if the trajectory ${\bf{h}}_{\boldsymbol{\theta }}^*\left(t\right)$ in which 
\begin{equation}
{\bf{P}} = \sum\nolimits_{k = 1}^{{k_c}} {{\theta _k}{{\bf{P}}_k}}  \in {\rm{conv}}\left( {\left\{ {{{\bf{P}}_k}} \right\}} \right), {\bf{t}} = {\bf{x}}_{C{R_\theta }}^*\left( \boldsymbol{\theta}  \right),
\label{equ:traj-robot}
\end{equation}
the trajectory ${\bf{h}}_{\boldsymbol{\theta }}^*\left(t\right)$ is optimal.
\subsection{Complexity Analysis}
In this subsection, the computational complexity of the optimization-based planning method is first analyzed, followed by that of the proposed method. Subsequently, a comparative analysis is conducted to highlight the differences in computational complexity between the two methods.

The optimization-based trajectory planning method formulates an optimization problem by treating trajectory parameters as optimization variables and solving for the optimal trajectory. Accordingly, the computational complexity of generating a single trajectory is $O(n^3_t)$, and the computational complexity of generating $k$ trajectories is $O(kn^3_t)$, where $n_t$ denotes the number of optimization variables.

\textcolor{blue}{The computational complexity of the trajectory planning method proposed in this paper can be divided into two parts: critical region generation and trajectory generation. As shown in \emph{Algorithm \ref{alg:paramprog}}, the critical regions are generated recursively, so its computational complexity depends not only on the number of optimization variables $n_t$ but also on the number of boundary trajectories $k_c$ and the number of regions $\lambda(\epsilon)$ resulting from the partitioning of the critical regions \cite{bemporad2006algorithm}. Here, $\lambda(\epsilon)$ is related to the error bound $\epsilon$ in the algorithm: a smaller $\epsilon$ leads to a more refined partition result and thus a larger $\lambda(\epsilon)$, whereas a larger $\epsilon$ results in a smaller $\lambda(\epsilon)$. Furthermore, it should be emphasized that $\lambda(\epsilon)$ does not increase indefinitely as $\epsilon$ decreases; it ceases to increase once the partition achieves perfect fitting (i.e., the exact optimal solution regions without approximation error).
Consequently, the computational complexity of critical region generation can be expressed as $O(\lambda(\epsilon)(n_t+k_c)^3)$.
For trajectory generation, the process involves first locating the corresponding regions within the critical regions and then generating the $k$ trajectory through affine functions. Thus, the computational complexity of this process is $O(kn_t)$, where $n_t$ denotes the number of optimization variables.
In summary, the total computational complexity of the proposed method for generating $k$ trajectories can be expressed as $O(\lambda(\epsilon)(n_t+k_c)^3 + kn_t)$,
where $n_t$ is the number of optimization variables and $k_c$ is the number of boundary trajectories with respect to $\boldsymbol{\theta}$ for multi-parametric programming.}

Compared to the computational complexity of the optimization-based planning method, the proposed method introduces the additional complexity of critical region generation. However, the computational complexity of trajectory generation is reduced from $O(n_t^3)$ to $O(n_t)$. \textcolor{blue}{More interestingly, the computational complexity of critical region generation $O(\lambda(\epsilon)(n_t+k_c)^3)$ is independent of the number of trajectories $k$.} This leads to the result that the number of trajectories generated has a negligible impact on the computational cost of the proposed method. \textcolor{blue}{Specifically, the proposed method exhibits an overall lower computational complexity than the traditional optimization-based method when the number of generated trajectories $k$ satisfies:
\begin{equation}
	k > \frac{\lambda(\epsilon)(n_t+k_c)^3}{n_t^3 - n_t}.
	\label{equ:complexity-condition}
\end{equation}
Since $k_c$ is typically small compared to $n_t$, the right side of (\ref{equ:complexity-condition}) can be approximated as $\lambda(\epsilon)$. This implies that when the number of desired trajectories $k$ exceeds the number of critical regions $\lambda(\epsilon)$ (i.e., $k \gtrsim \lambda(\epsilon)$), the proposed method demands less computational load. Moreover, generating a massive number of trajectories is a highly common and easily attainable requirement in practical swarm trajectory planning scenarios.
As $k$ exceeds this threshold and further increases, the computational advantage of the proposed method becomes more pronounced.}

\section{Controller}
\begin{figure}
	\centering
	\includegraphics{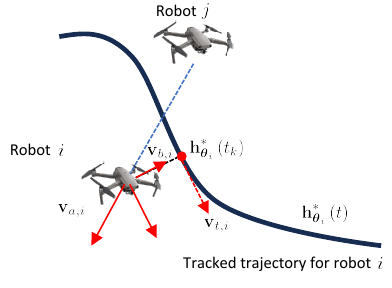}
	\caption{The controller for the robot $i$ in the swarm. The black curve represents the optimal trajectory tracked by the robot $i$. The red point denotes the corresponding point ${\bf{h}}_{\boldsymbol{\theta }_i}^*\left(t_k\right)$ in the trajectory ${\bf{h}}_{\boldsymbol{\theta }_i}^*\left(t\right)$ to the robot $i$. And, the three vectors represent feedforward command ${{\bf{v}}_{t,i}}$, tracking command ${{\bf{v}}_{b,i}}$, and avoidance command ${{\bf{v}}_{a,i}}$ respectively, in which the feedforward command ${{\bf{v}}_{t,i}}$ is the tangent vector of the point ${\bf{h}}_{\boldsymbol{\theta }_i}^*\left(t_k\right)$ and the avoidance command ${{\bf{v}}_{a,i}}$ is align with the line between robot $i$ and robot $j$.}
	\label{fig:controller}
\end{figure}
The controller combining trajectory tracking and collision avoidance is described in the following.
Suppose the parameters of the trajectory for the robot $i$ to track are ${\boldsymbol{ \theta}}_i$. Then, according to (\ref{equ:traj-robot}) and (\ref{equ:bezier-mat}), the optimal trajectory ${\bf{h}}_{\boldsymbol{\theta }_i}^*\left(t\right)$ is generated by affine functions which have a low computational complexity.

The controller shown in Fig. \ref{fig:controller} refers to the work in virtual tube control problem \cite{Quan2021Distributed}, which can be expressed as
\[{{\bf{v}}_{c,i}} = {{\bf{v}}_{t,i}} + {{\bf{v}}_{b,i}} + {{\bf{v}}_{a,i}},\]
where ${\bf v}_{c,i}$ is the velocity command for robot $i$ in (\ref{equ:robotmodel}), ${{\bf{v}}_{t,i}} = \frac{{{\rm{d}}{\bf{h}}_{{{\boldsymbol{\theta }}_i}}^*\left( t \right)}}{{{\rm{d}}t}}$ is the feedforward velocity from the trajectory, ${{\bf{v}}_{b,i}}$ is the command to track the trajectory and ${{\bf{v}}_{a,i}}$ is the avoidance command to avoid the collision with other robots. 
By integrating feedforward, tracking, and avoidance commands, the controller enables the robot to track the planned trajectory while ensuring safety and avoiding collisions.

\begin{rmk}
	The details of the stability proof for the controller, as well as the specific formulations of ${\bf v}_{b,i}$ and ${\bf v}_{a,i}$, can be found in the relevant proof in \cite{Quan2021Distributed}.
\end{rmk}
\section{Simulations and Experiments}
In this section, the theoretical effectiveness of the proposed method is first validated through numerical experiments. Then, its advantages in flight safety and efficiency are demonstrated through flight simulations. 
\textcolor{blue}{Next, hardware-in-the-loop (HIL) simulations are performed to compare the proposed method with SOTA methods including EGO-swarm \cite{zhou_ego-swarm_2021}, AMSwarmX \cite{adajania2024amswarmx}, and RMADER \cite{kondo2023robust} }showcasing the computational efficiency and real-time performance of the proposed method.
Finally, swarm flight experiments in unknown environments are implemented to evaluate and validate the real-time performance and effectiveness of the proposed method. 
\subsection{Simulations and Experiments Setup}
\begin{figure}
	\centering
	\includegraphics[width=0.7\linewidth]{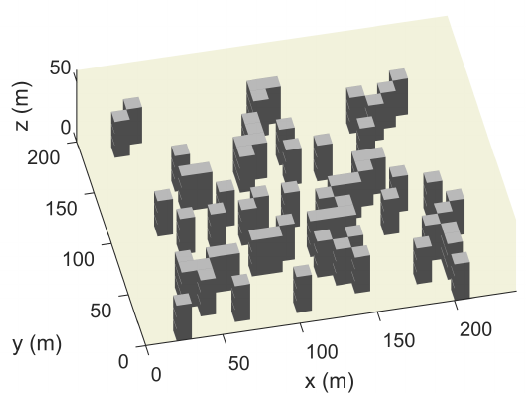}
	\caption{The 3-D environment with random obstacles used for comparative validations.}
	\label{fig:matlab-env}
\end{figure}

The numerical experiments for theoretical validation and flight simulations are conducted on a PC with a CPU i9-13900KF and 32GB RAM, using MATLAB for coding. The 3-D obstacle environment, shown in Fig. \ref{fig:matlab-env}, is 250$\times$200$\times$30m, and the size of the random obstacles is 10$\times$10$\times$30m.

\begin{figure}
	\centering
	\includegraphics[width=0.9\linewidth]{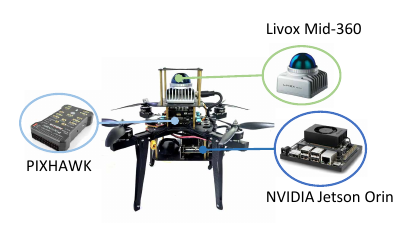}
	\caption{The drone is used for HIL simulations and real flight experiments.}
	\label{fig:dronepic}
\end{figure}

\begin{figure}
	\centering
	\includegraphics[width=\linewidth]{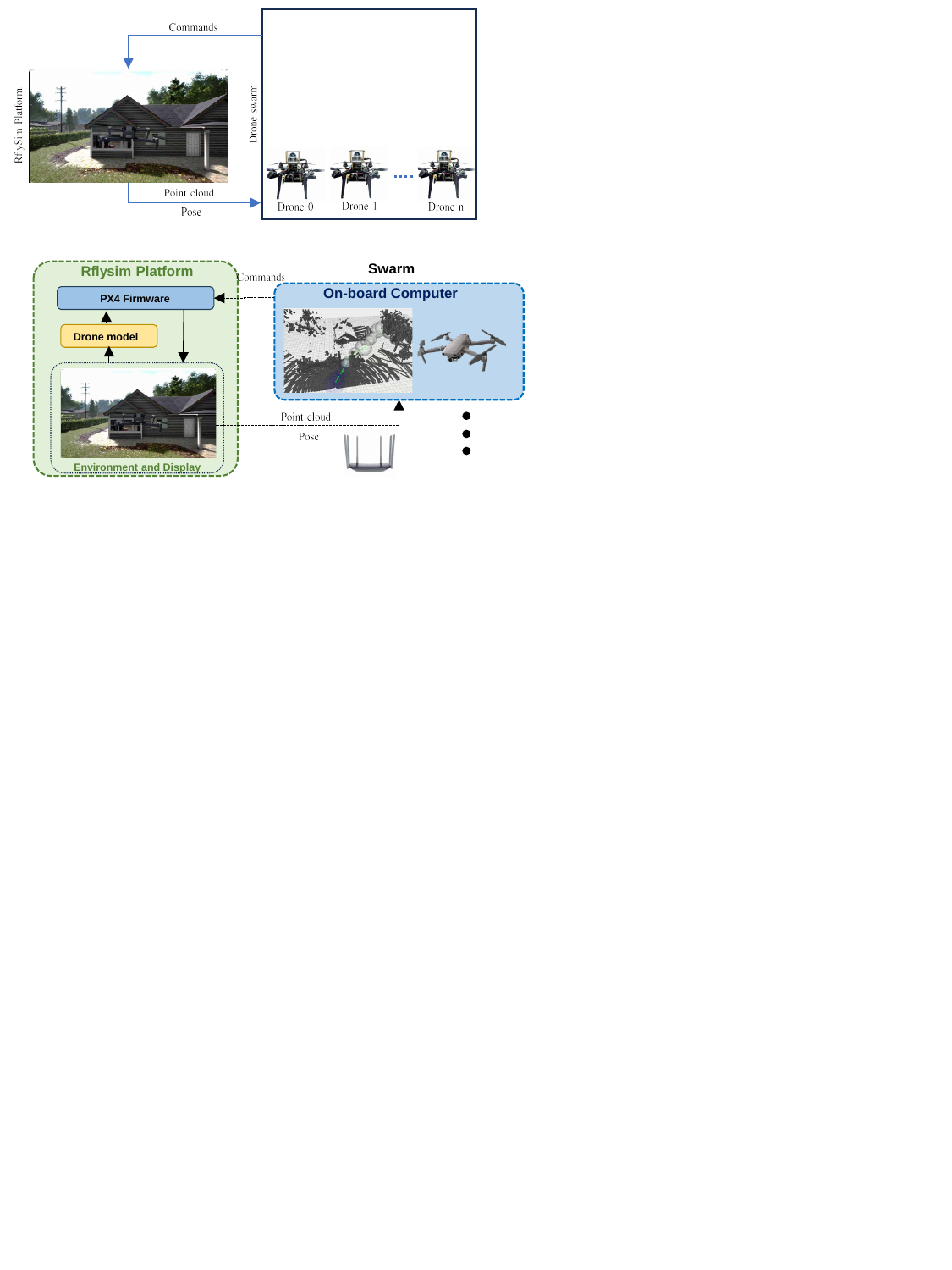}
	\caption{The hard-in-the-loop (HIL) simulation platform.}
	\label{fig:hitl}
\end{figure}

For HIL simulation to verify the real-time performance and the computational efficiency of the proposed method, the platform includes drones shown in Fig. \ref{fig:dronepic} and RflySim \cite{DAI2021106727} simulator shown in Fig \ref{fig:hitl}. The drone is equipped with a LiDAR (MID-360), a flight controller, and an onboard computer (NVIDIA Jetson Orin). The code is written in C++ and runs on the Ubuntu 20.04 OS. The solver used in the proposed method is MOSEK. 

\begin{figure}
	\centering
	\includegraphics{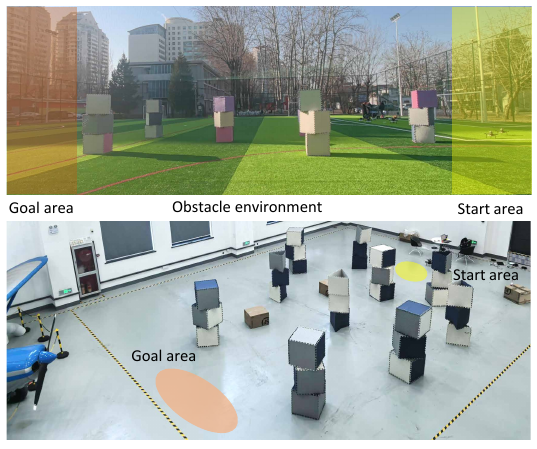}
	\caption{The unknown environments with random obstacles. The \textcolor{blue}{yellow and pink }blocks represent the start and goal areas, respectively.}
	\label{fig:env1}
\end{figure}

For the real flight experiments, drones identical to those used in the HIL simulations are employed. The LiDAR is used for autonomous localization and generates the point cloud map for swarm trajectory planning and control. Inter-drone communication is implemented via WiFi.
The unknown obstacle environments are shown in Fig. \ref{fig:env1}, with obstacles randomly placed. The drone swarm takes off from the start area, navigates through the obstacle environment, and reaches the goal area. The safety distance between drones is 0.8m. The number of drones in the swarm is 3.
\subsection{Numerical Validation}
In this subsection, the process of optimal virtual tube planning is briefly described. Subsequently, the theoretical effectiveness of the proposed planning method is validated, while the effects of the error bound $\epsilon$ of the proposed method on the approximation results are investigated.

\begin{figure*}
	\centering
	\includegraphics{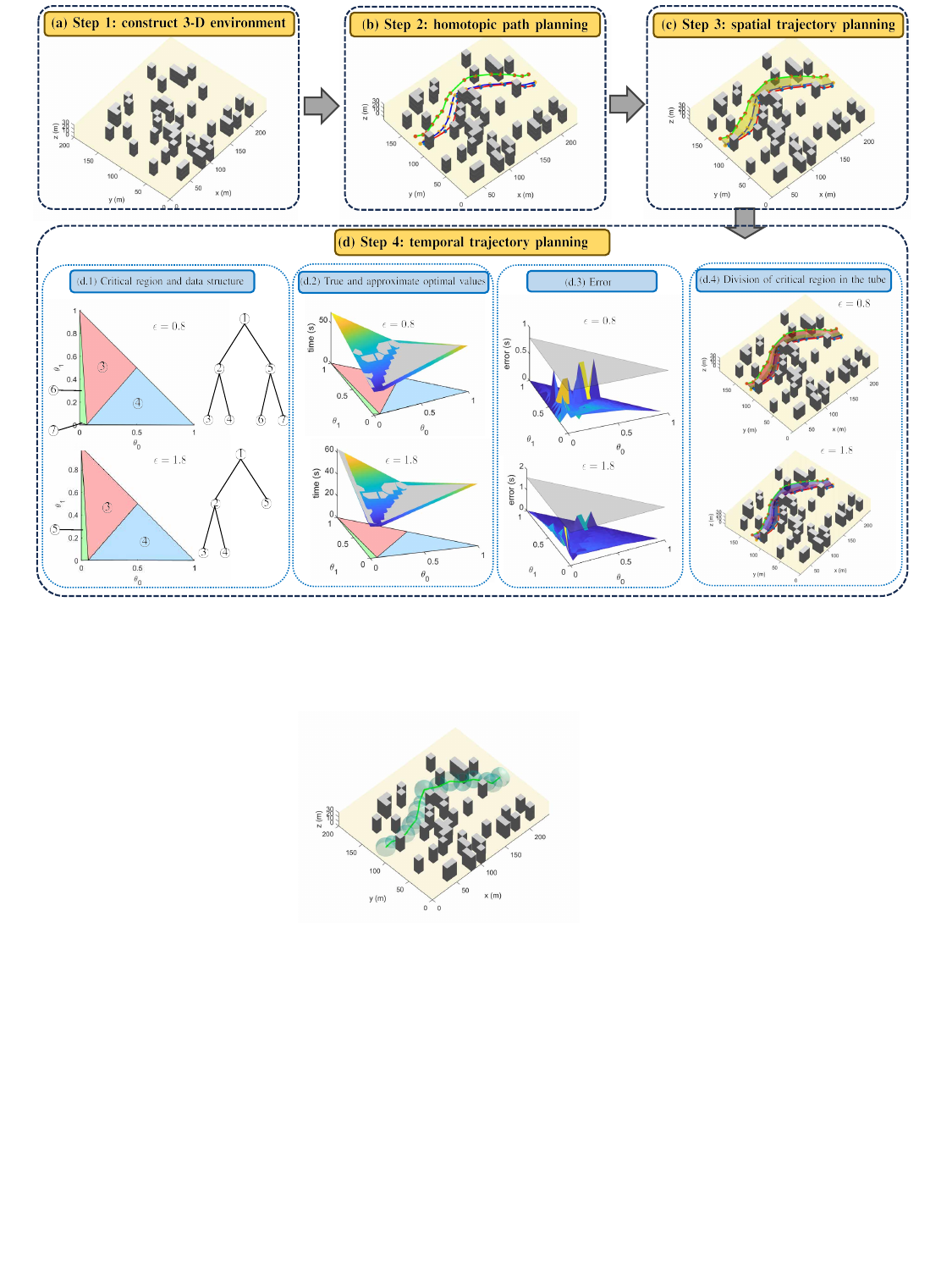}
	\caption{The process and validation of optimal virtual tube planning. (a) A random obstacle environment in 3-D space is constructed. (b) The homotopic paths denoted by the colorful curves are generated by the Tube RRT* algorithm. (c) The infinite set of spatial trajectories is represented by the yellow area among the boundary trajectories. \textcolor{blue}{(d) Temporal trajectory planning is performed based on the spatial trajectory planning. (d.1) Algorithm 1, according to different error bounds $\epsilon$, finds and computes the parameter values for a varying number of critical trajectories. When mapped into the parameter region, these critical trajectories act as vertices that partition the parameter region into different critical regions. The different colors of triangles represent the different regions in critical regions. Since the parameter domain is iteratively subdivided by the algorithm, storing the critical region in a tree structure is most efficient for querying. Note that the numbering of critical regions does not start from 1, whereas the tree structure includes all numbering starting from 1. This reflects the iterative algorithm: the final critical regions correspond to the leaf nodes of the tree structure, meaning each is a further subdivision of the critical regions generated in the previous iteration.
	(d.2) The approximate and true values generated for critical regions. The gray surfaces represent the approximate optimal values by multi-parametric programming, and the colorful surfaces represent the true optimal values by trajectory planning. (d.3) The error between the approximate and true values for the critical regions.} The gray surface represents the error bound $\epsilon$, and the colorful surface represents the error between the approximate and true optimal values. (d.4) The different colors represent the different regions divided by the critical regions in the optimal virtual tube.}
	\label{fig:simvalidation}
\end{figure*}

The 3-D obstacle environment is shown in Fig. \ref{fig:simvalidation}(a). With fixed start and goal regions, the homotopic paths in the same homotopy class are planned by \cite{mao2024tube}, as depicted in Fig. \ref{fig:simvalidation}(b). Based on the homotopic paths, the optimal virtual tube planning is then divided into two steps: spatial trajectory planning and temporal trajectory planning. The step of spatial trajectory planning, illustrated in Fig. \ref{fig:simvalidation}(c), generates an infinite set of optimal trajectories with initial time allocations. The step of temporal trajectory planning, shown in Fig. \ref{fig:simvalidation}(d), generates the optimal time allocations for all trajectories in the infinite set of trajectories.   

In Step 3, for any trajectory in the infinite set of spatial trajectories, the control points can be approximated by the convex combination of control points of boundary trajectories \cite{mao2023optimal}. However, in Step 3, all trajectories use the same initial time allocations. Therefore, in Step 4, the time allocations for all trajectories are optimized. Here, the critical regions are generated by \emph{Algorithm \ref{alg:paramprog}}, as shown in Fig. \ref{fig:simvalidation}(d.1). Any trajectory can be mapped to a specific region within the critical regions, and the optimal time allocations for that trajectory can be linearly approximated using the three vertices of the corresponding region.
To numerically validate this method, hundreds of points are uniformly sampled within the critical regions. The true optimal values and their approximations are then computed under different error bounds $\epsilon$, as shown in Fig. \ref{fig:simvalidation}(d.2). As demonstrated in Fig. \ref{fig:simvalidation}(d.3), the proposed method successfully approximates the optimal values within the specified error bound $\epsilon$. For a more intuitive representation, the critical regions are visually mapped onto the trajectories, as illustrated in Fig. \ref{fig:simvalidation}(d.4).


\subsection{Flight Simulations}
In this subsection, the performance of the drone swarm within the optimal virtual tube  is validated in the same environment. The obstacle environment and the parameter settings of the optimal virtual tube are consistent with those in the previous subsection.
The drone swarm consists of 36 drones, with the radius \( r_s \) of the safety area set to 0.4m and the radius $r_a$ of the avoidance area set to 1m.
The time trajectory planning uses the results from the initial time interval as well as $\epsilon = 0.8$. 

\begin{table*}[]
	\begin{tabular}{|l|ll|ll|ll|ll|}
		\hline
		\multirow{2}{*}{Number of obstacles} & \multicolumn{2}{c|}{20}                       & \multicolumn{2}{c|}{50}                       & \multicolumn{2}{c|}{70}                       & \multicolumn{2}{c|}{100}                       \\ \cline{2-9} 
		& \multicolumn{1}{l|}{Mean speed} & Flight time & \multicolumn{1}{l|}{Mean speed} & Flight time & \multicolumn{1}{l|}{Mean speed} & Flight time & \multicolumn{1}{l|}{Mean speed} & Flight time \\ \hline
		Initial time allocation              & \multicolumn{1}{l|}{4.86 m/s}           &    36.79 s         & \multicolumn{1}{l|}{4.21 m/s}           &53.85s             & \multicolumn{1}{l|}{4.03 m/s}           & 54.31 s            & \multicolumn{1}{l|}{3.35 m/s}           & 55.35 s            \\ \hline
		Approximate time allocation          & \multicolumn{1}{l|}{5.58 m/s}           & 32.48 s             & \multicolumn{1}{l|}{4.61 m/s}           &34.99s             & \multicolumn{1}{l|}{4.52 m/s}           & 35.67 s            & \multicolumn{1}{l|}{4.95 m/s}           & 37.10 s            \\ \hline
	\end{tabular}
	\caption{\textcolor{blue}{The mean speed and flight time of the swarm with different time allocations under different numbers of obstacles.}}
	\label{table:2}
\end{table*}

\begin{figure}
	\centering
	\includegraphics{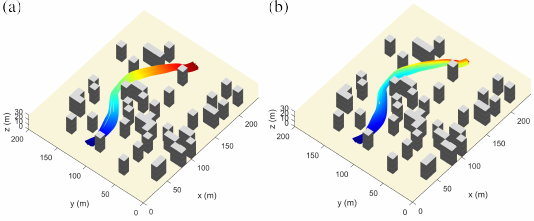}
	\caption{The trajectories of swarm with different time allocations. The different colors of the trajectories correspond to different time stamps. (a) Results for initial time allocations. (b) Approximate time allocations with $\epsilon=0.8$.}
	\label{fig:flightsim-traj}
\end{figure}

\begin{figure}
	\centering
	\includegraphics{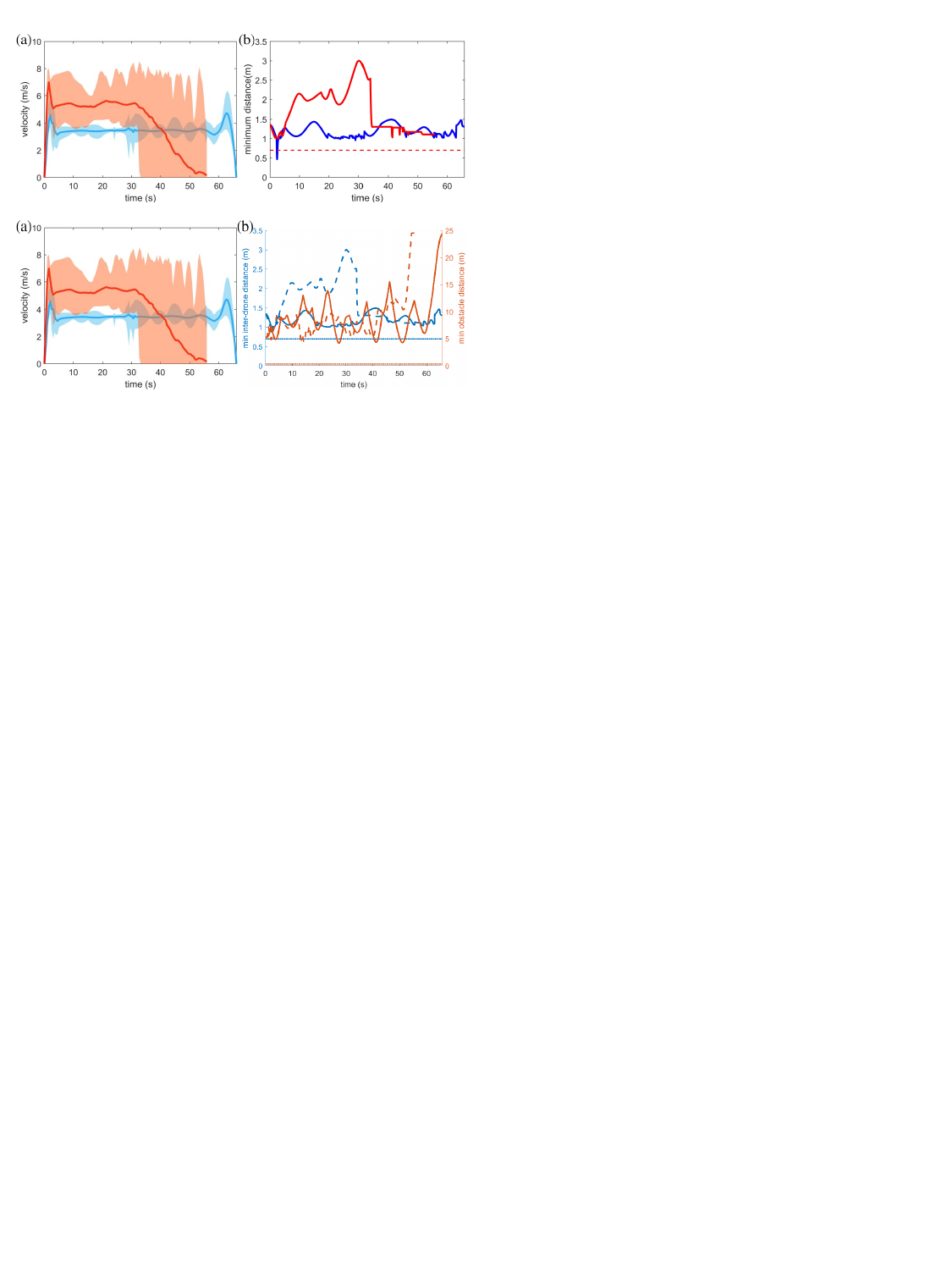}
	\caption{Speed distribution and the minimum distance among drones. (a) Speed distribution of the swarm: The orange and blue areas show the speed range for $\epsilon=0.8$ and the initial time allocations, while the red and blue curves represent the corresponding average speed. (b) \textcolor{blue}{ Minimum distances during flight simulation. The left axis (blue) shows the minimum inter-drone distance, and the right axis (orange) shows the minimum distance to obstacles. Solid and dashed lines correspond to initial and $\epsilon=0.8$ time allocations, respectively. The dotted lines indicate safety thresholds: 0.7 m for inter-drone separation and 0.35 m for obstacle clearance. All distances remain above their respective safety bounds throughout the flight.}}
	\label{fig:flightsim-speed-dis}
\end{figure}

The trajectory results of the swarm are shown in Fig. \ref{fig:flightsim-traj}, where different colors along the trajectories represent different time stamps. A comparison of the trajectories reveals differences in time allocations between the various optimal trajectories. 
As shown in Fig. \ref{fig:flightsim-speed-dis}(a), the swarm with approximate temporal trajectory planning reaches the goal area faster and in a shorter time. This observation is further corroborated by the speed and inter-drone distance data, as shown in Fig. \ref{fig:flightsim-speed-dis}. 
From Fig. \ref{fig:flightsim-speed-dis}(a), it can be observed that the swarm with approximate temporal trajectory planning reaches the goal area faster and in a shorter time. 
In the initial time allocation, the high number of drones in the swarm leads to unsafe distances in narrower areas, as depicted in Fig. \ref{fig:flightsim-speed-dis}(b). Additionally, the inter-drone safety distance is larger when the swarm is more dispersed, in contrast to the swarm with the initial time allocations. This further emphasizes the influence of temporal trajectory planning on both swarm speed and safety.
\textcolor{blue}{To further evaluate the effectiveness of the proposed temporal trajectory planning, an ablation study is conducted. As summarized in Table \ref{table:2}, environments with varying obstacle densities, explicitly containing 20, 50, 70, and 100 obstacles respectively, are tested. To ensure the reliability and statistical significance of the results, multiple sets of random testing environments are generated and evaluated for each specified number of obstacles. The table compares the performance metrics of the drone swarm utilizing the initial time allocation against the approximate time allocation. The experimental results explicitly indicate that, across all obstacle configurations, employing the approximate time allocation consistently achieves a higher mean speed and a significantly reduced total flight time. This ablation study robustly demonstrates the overall advantage and effectiveness of the proposed temporal trajectory planning method within various complex settings.}


\subsection{Numerical Analysis of Computation Time}
\textcolor{blue}{The efficient computation time of the proposed method is compared with EGO-swarm, AMSwarmX, and RMADER from two aspects: one is the effect of different replanning frequencies on the computational time, and the other is the effect of the number of trajectories on the total computational time.} All simulations are conducted on the hardware platform to maintain consistency with the real-world experiments. 
Meanwhile, due to the differences between the framework of the proposed method and those of EGO-swarm, AMSwarmX, and RMADER, two metrics are adopted for comparison: individual computation time and total computation time. The former refers to the computation time required for each robot, while the latter denotes the sum of the computation times of all robots in the swarm.

To analyze the effect of replanning frequency on computation time, the HIL simulations with three drones are implemented on the RflySim platform. The same start and goal areas are set where the drones navigate through a forested environment. The computation times of two methods in different frequencies are listed in Table \ref{table:1}. 
In Table \ref{table:1}, since the proposed method is a centralized trajectory planning approach, it is necessary to distinguish between the individual computation time of the leader drone and that of the other drones. From the computation times in Table \ref{table:1}, it can be observed that as the replanning frequency increases, the computation times for both methods increase. Although the individual computation time for the leader drone in our method is consistently higher, the total computation time remains smaller due to the very low computation times of the other drones. Additionally, it can be seen that, due to the minimal computation time for the calculations, the effect of replanning frequency on the individual computation time of the other drones can be considered negligible.

\begin{figure}
	\centering
	\includegraphics[width=0.8\linewidth]{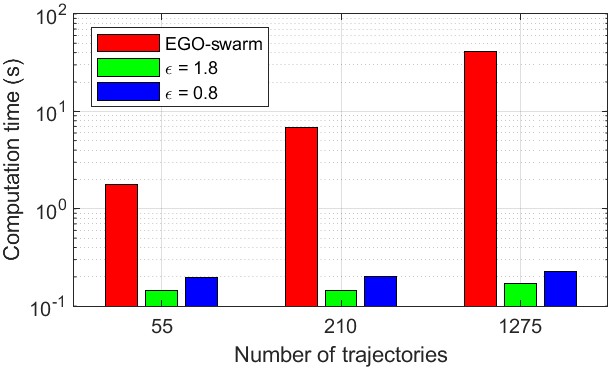}
	\caption{Computation time vs Number of trajectories. The red, green, and blue blocks represent the total computation times of \textcolor{blue}{EGO-swarm}, $\epsilon=1.8$, and $\epsilon=0.8$ respectively. }
	\label{fig:computation-time}
\end{figure}

\textcolor{blue}{
\begin{table*}[]
	\begin{center}
		\begin{tabular}{|cl|c|c|c|c|}
			\hline
			\multicolumn{2}{|l|}{Replan Frequency (Hz)} & 1                & 2                & 5                & 10               \\ \hline
			\multicolumn{1}{|c|}{\multirow{4}{*}{Computation time (s)}} & EGO-swarm & -\textbar0.009\textbar0.029 & -\textbar0.013\textbar0.037 & -\textbar0.031\textbar0.094 & -\textbar0.047\textbar0.141 \\ \cline{2-6} 
			\multicolumn{1}{|c|}{}      & Ours          & 0.015\textbar0.92$\times10^{-6}$\textbar0.015 & 0.021\textbar2.64$\times10^{-6}$\textbar0.021 & 0.082\textbar5.83$\times10^{-6}$\textbar0.082 & 0.117\textbar1.29$\times10^{-6}$\textbar0.117 \\ \cline{2-6} 
			\multicolumn{1}{|c|}{}      & AMSwarmX      & -\textbar0.011\textbar0.038    & -\textbar0.018\textbar0.045    & -\textbar0.052\textbar0.101    & -\textbar0.068\textbar0.285    \\ \cline{2-6} 
			\multicolumn{1}{|c|}{}      & RMADER        & -\textbar0.013\textbar0.041    & -\textbar0.020\textbar0.047    & -\textbar0.058\textbar0.110    & -\textbar0.072\textbar0.301    \\ \hline
		\end{tabular}
		\caption{\textcolor{blue}{Comparison of computation time at different frequencies: EGO-swarm\cite{zhou_ego-swarm_2021}, AMSwarmX \cite{adajania2024amswarmx},  RMADER \cite{kondo2023robust} versus Ours. The individual computation time for the leader drone and other drones and the total computation time at each frequency for each method are separated by vertical bars.}}
		\label{table:1}
	\end{center}
\end{table*}
}
The total computation times for the optimization-based baseline methods and the proposed method, under varying numbers of trajectories and different error bounds $\epsilon$, are presented in Fig. \ref{fig:computation-time}. \textcolor{blue}{ Since EGO-swarm, AMSwarmX, and RMADER are all optimization-based methods, they exhibit the same qualitative trend with respect to the number of generated trajectories. Therefore, EGO-swarm is selected here as a representative optimization-based baseline for comparison.} As observed, the computation time of the optimization-based baseline represented by EGO-swarm increases with the number of trajectories, while the proposed method is less sensitive to this factor, exhibiting only a slight increase. This minor increase can be attributed to the time required for linear computations, which is significantly shorter than the time needed for optimization solving. 
\textcolor{blue}{Furthermore, the computation time is influenced by the size of the error bound \(\epsilon\): \(\epsilon\) essentially affects the number of critical regions \(\lambda\) in the computational complexity \(O\big(\lambda (n_t + k_c)^3\big)\); a smaller \(\epsilon\) requires higher approximation accuracy, which increases \(\lambda\) (i.e., demands finer partitioning of critical regions) and therefore leads to longer total computation time. }
Overall, the proposed method achieves an order-of-magnitude reduction in computation time compared to the optimization-based baseline represented by EGO-swarm, remains minimally affected by the number of trajectories, and requires a chosen error bound to balance computational cost and accuracy. \textcolor{blue}{These empirical results are consistent with the complexity analysis in Section V-C, further validating the efficiency of the proposed method.}

\subsection{Experiment Results}
In this subsection, several experiments are implemented to evaluate the proposed method. Initially, comparative experiments are conducted to analyze the effect of obstacle positioning in environments on swarm flight. 
\textcolor{blue}{Then, two real-flight scenarios—an indoor scenario (Fig. \ref{fig:real-flight-2}(a)) and an outdoor scenario (Fig. \ref{fig:real-flight-1}(a))—were used to validate the real-time performance of the proposed method. The hardware platform and sensor suite were identical in both experiments; the primary differences arose from environmental disturbances and obstacle density. The indoor tests were performed in a confined space, enabling high-density obstacle scenarios and facilitating controlled comparative experiments. The outdoor tests were conducted in a larger, open environment, where obstacle distributions were typically sparser. These complementary indoor and outdoor experiments demonstrate the proposed method’s effectiveness across a range of operating conditions.}

\subsubsection{Comparative Experiments}
\begin{figure*}
	\centering
	\includegraphics{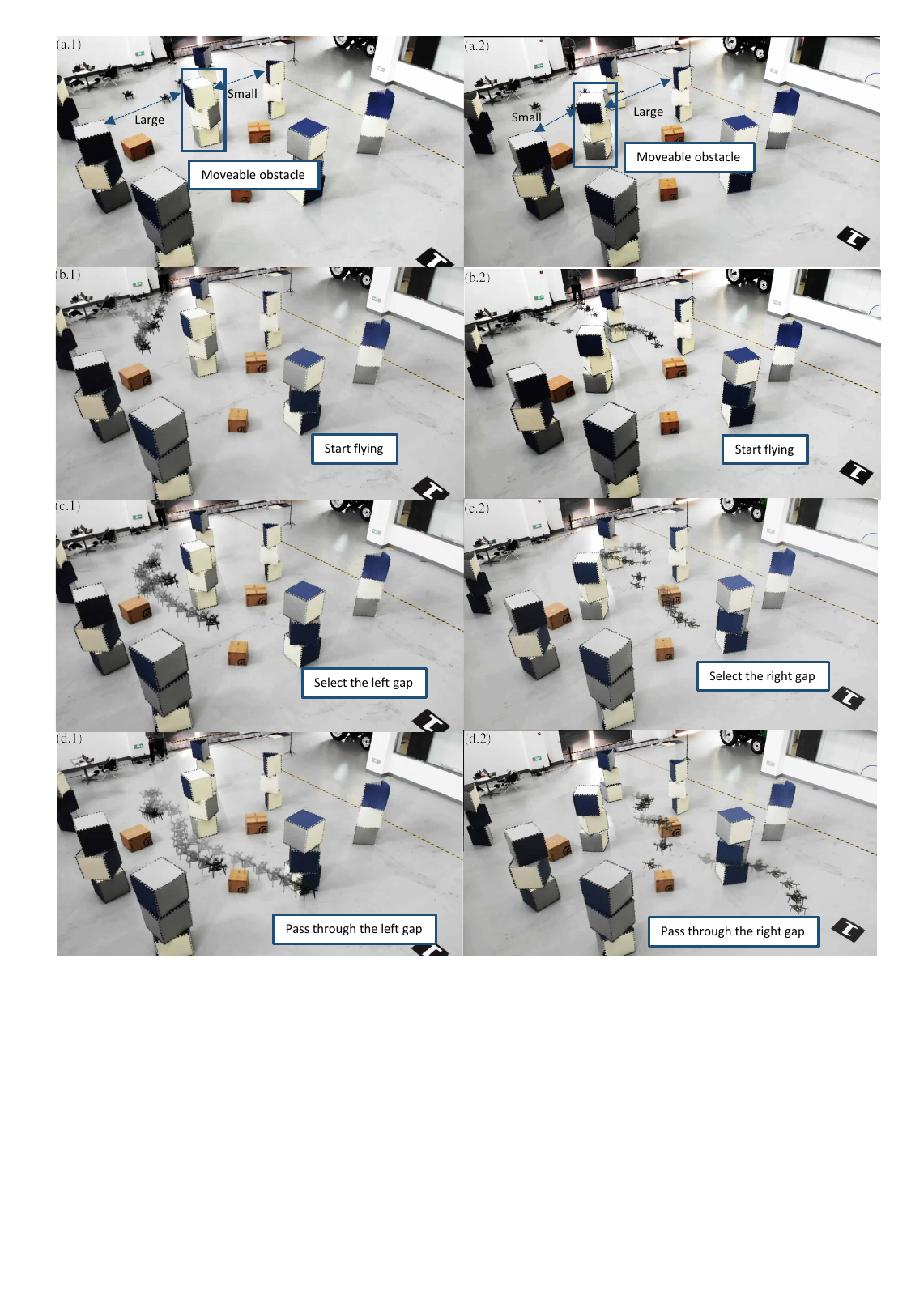}
	\caption{Comparative experiments in scenario 1 with a moveable obstacle. (a) Obstacle at different positions: in (a.1), the large gap is on the left; in (a.2), the large gap is on the right. (b) Trajectories of the swarm at the start of flight in the respective environments.
		(c) Trajectories of the swarm as they approach and select the large gap.
		(d) Trajectories of the swarm passing through the large gap.}
	\label{fig:comparative-experiments}
\end{figure*}

The experimental configuration comprises a priori unknown indoor environment, as illustrated in Fig. \ref{fig:comparative-experiments}(a). \textcolor{blue}{The takeoff configuration was kept fixed across all experiments, with robot 0 maintained as the leader robot throughout.} Obstacle spatial constraints were manipulated by dynamically adjusting a moveable obstacle to modulate the gap size. A comparative experiment was conducted pre- and post-obstacle displacement, with the results of swarm trajectories depicted in Fig. \ref{fig:comparative-experiments}(b)-(d).
Experimental results demonstrate that the centralized trajectory planning framework induces homotopy-equivalent swarm movement patterns, while the implemented homotopy-based path planning algorithm achieves to select the larger gap within the obstacle environments.

\subsubsection{Indoor Experiment}
\begin{figure*}
	\centering
	\includegraphics{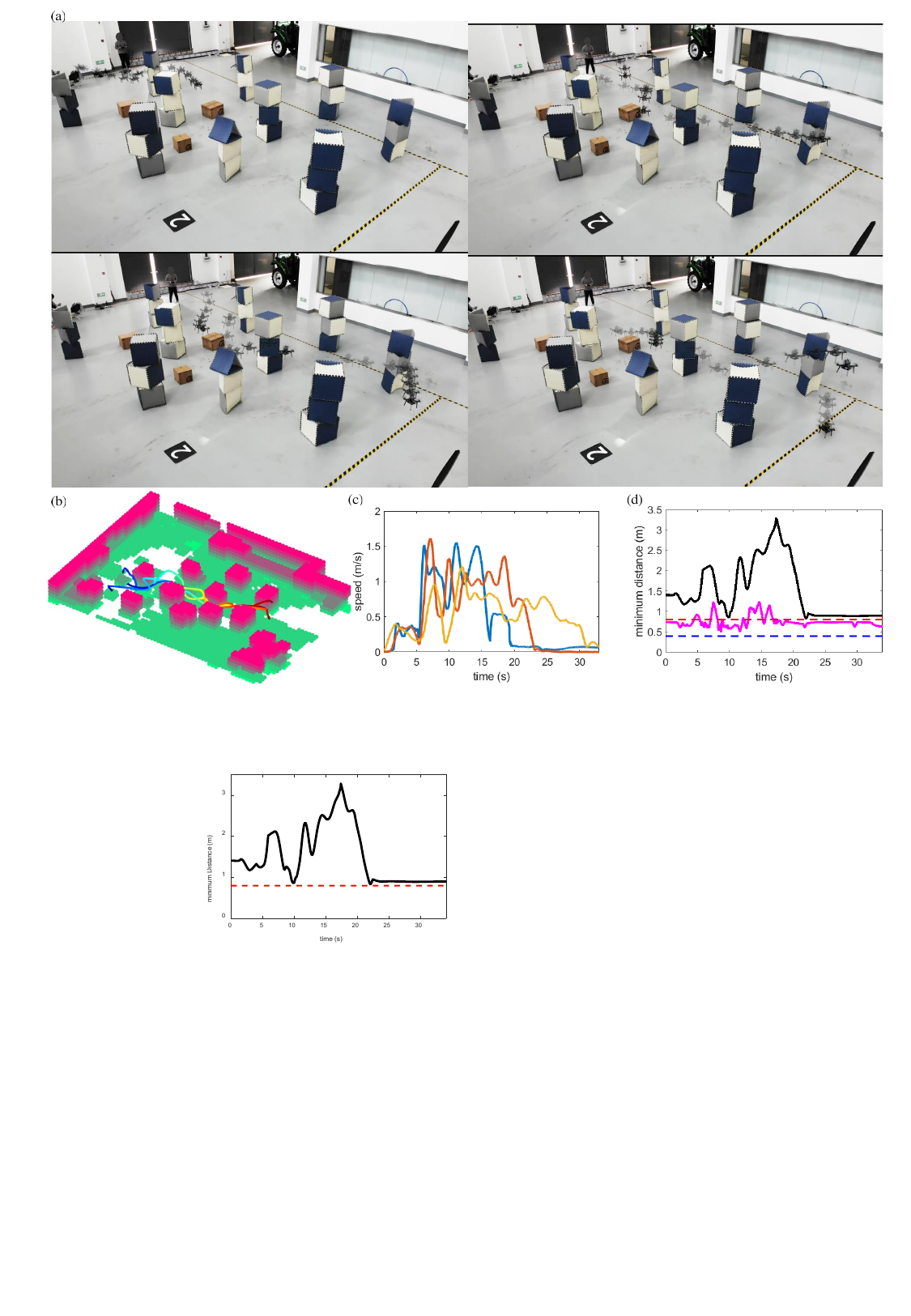}
	\caption{The real flight experiments in scenario 2. (a) The composite image of drone swarm flight. (b) The swarm trajectories in the point cloud maps. The trajectories at different time stamps are represented in different colors. (c) Swarm speeds over time. The red, yellow, and blue curves represent the speed of drones 0, 1, and 2, respectively. (d) \textcolor{blue}{The minimum inter-drone distance and the minimum drone-obstacle distance. The red and blue dotted lines represent the safety distances for inter-drone separation and drone-obstacle separation, respectively. The black and pink curves denote the minimum inter-drone distance and the minimum drone-obstacle distance, respectively.}}
	\label{fig:real-flight-2}
\end{figure*}

For the indoor environment with dense obstacles, the time-varying trajectories, shown in Fig. \ref{fig:real-flight-2}(a)-(b), demonstrate that temporal trajectory planning enables the swarm to stagger their passage through narrow gaps, effectively reducing the risk of collisions. As presented in Fig. \ref{fig:real-flight-2}(c), the maximum flight speed in the dense obstacle environment reached 1.6 m/s, while the minimum inter-drone distance remained above the safety distance of 0.8 m.
\subsubsection{Outdoor Experiment}
\begin{figure*}
	\centering
	\includegraphics{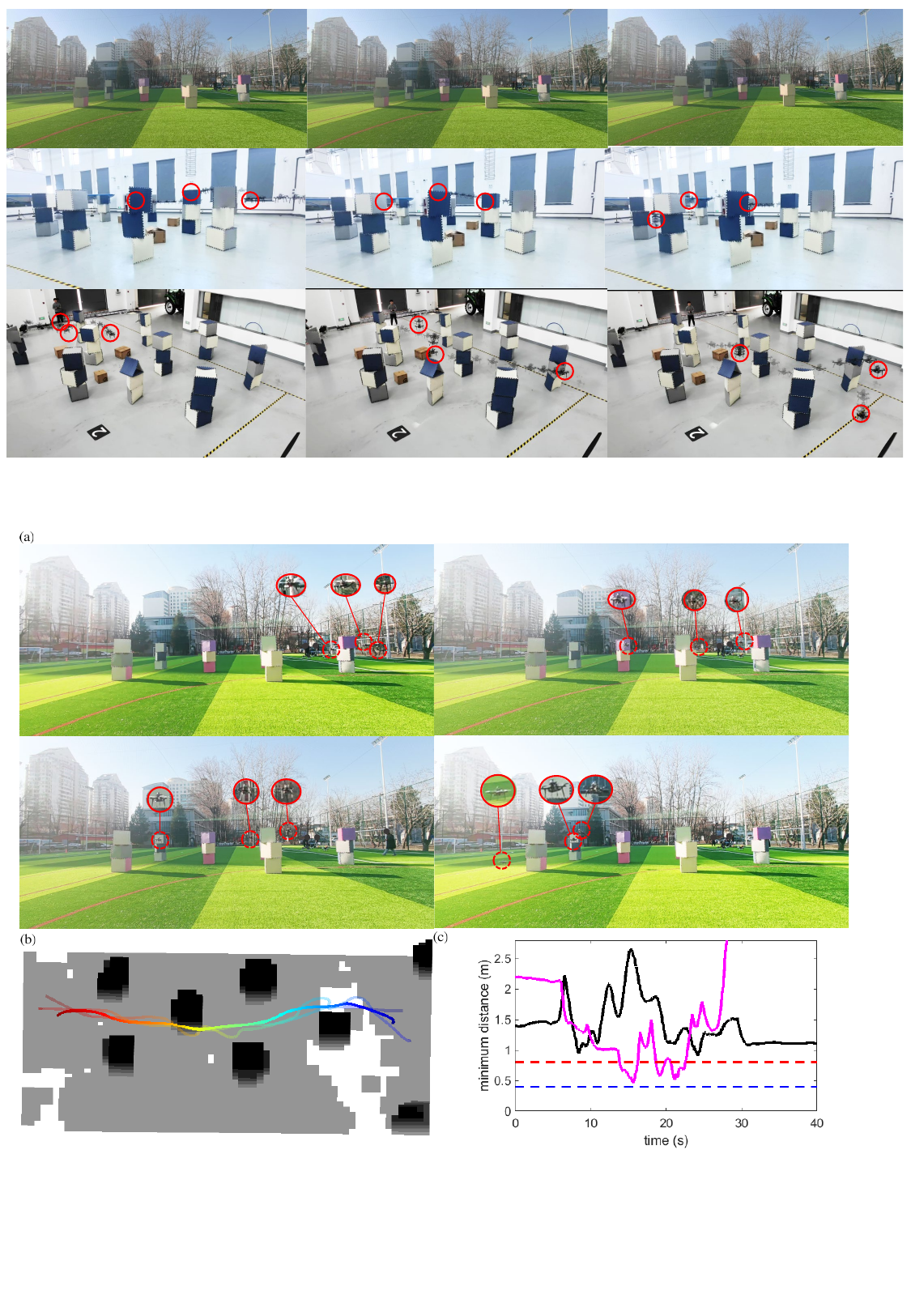}
	\caption{The real flight experiments in scenarios 3. (a) The composite image of drone swarm flight. (b) The swarm trajectories in the point cloud maps. \textcolor{blue}{Trajectories at different time stamps are color-coded; the trajectory of robot 1 is plotted as a solid (opaque) line, while the trajectories of the other drones are shown as semi-transparent lines.} (c) \textcolor{blue}{The minimum inter-drone distance and the minimum drone-obstacle distance. The red and blue dotted lines represent the safety distances for inter-drone separation and drone-obstacle separation, respectively. The black and pink curves denote the minimum inter-drone distance and the minimum drone-obstacle distance, respectively.}}
	\label{fig:real-flight-1}
\end{figure*}

In the outdoor environment with sparse obstacles, the time-varying trajectories depicted in Fig. \ref{fig:real-flight-1}(a)-(b) illustrate a similar staggered passage strategy, ensuring safe navigation through obstacles. The swarm achieved a higher maximum speed of 2 m/s in the sparse obstacle environment, while maintaining a safe inter-drone distance of at least 0.8 m, as shown in Fig. \ref{fig:real-flight-1}(c).

These results, which closely align with simulation outcomes, further validate the real-time performance and effectiveness of the proposed method in real-world environments.


\subsection{Discussion}
Several key observations can be discussed based on the results from the simulations and the real flight experiments.
The proposed centralized trajectory planning method demonstrates advantages over traditional methods of efficient computation, homotopic trajectory planning, and collaboration within the swarm. These advantages not only enhance the performance of the swarm but also provide strong support for real-world applications, particularly in complex environments requiring coordinated tasks.
\subsubsection{Efficient Computation}
The proposed centralized trajectory planning method offers significant computational efficiency compared to traditional methods. Although there is a slight increase in the computation time for the leader drone, applying multi-parametric programming with a linear approximation of the optimal solution greatly reduces the computation time for the other drones. This method results in a substantial decrease in the total computation time of the swarm. As the number of drones in the swarm increases, the reduction in computation time becomes more pronounced. Specifically, the centralized method allows for collaborative computation within the swarm and the efficient allocation of resources, ensuring that each robot only performs key calculations at critical moments rather than redundantly solving optimization problems for every task. The proposed method not only improves computational efficiency but also ensures real-time responsiveness, meeting the requirements of practical applications.
\subsubsection{Homotopy Trajectories}
One key advantage of centralized trajectory planning is that it ensures all drones in the swarm follow trajectories within the same homotopy class. This means that all drones adhere to a unified trajectory planning pattern, ensuring that the trajectories of the swarm maintain topological consistency, unlike distributed trajectory planning where non-homotopic trajectories may emerge. In specific applications, such as the transportation of payloads for swarms in complex environments, it is critical for the swarm to maintain coordinated and consistent trajectories to avoid interference between drones. The centralized trajectory planning ensures that the trajectories are synchronized, allowing for more efficient task execution, particularly in tasks requiring precise coordination.
\subsubsection{Swarm Collaboration}
While a distributed planning framework does not require the division of tasks among the agents, a centralized planning framework, where each agent consumes different amounts of computational resources, necessitates a hierarchical framework of the swarm. Although the centralized planning framework makes the swarm structure more complex, it also optimizes the allocation of computational resources. For example, more computational resources can be allocated to the leader drone to handle more computations, while other drones can be tasked with specific duties such as swarm localization or sensing and recognition tasks. This hierarchical framework allows for efficient task execution, ensuring that the computational load of each agent is balanced and that the overall swarm operates optimally. By assigning resources based on the importance and complexity of tasks, the swarm can collaborate more efficiently and maximize the utilization of the overall computational resources.
\section{Conclusion and Future Work}
For autonomous navigation for \textcolor{blue}{robot swarms} in unknown environments, smooth motion with low computation cost is a significant challenge. This paper is based on multi-parametric programming to generate optimal trajectories with linear computation complexity, greatly enhancing re-planning frequency. Additionally, an update strategy is designed to enable real-time operation in unknown environments. The effectiveness and real-time performance of the method are validated through simulations and experiments.
In future work, \textcolor{blue}{distributed optimal virtual tube planning in communication-denied environments and large-scale experiments} will be investigated to improve the reliability of its applications.

%
%
%
%
\appendices

\bibliographystyle{IEEEtran} 
\bibliography{bib/IEEEfull,bib/IEEEbib}

\vfill

\end{document}